\documentclass{article}

\usepackage{microtype}
\usepackage{graphicx}
\usepackage{subcaption}
\usepackage{booktabs} 

\usepackage{hyperref}

\usepackage[preprint]{icml2026}

\usepackage{amsmath}
\usepackage{amssymb}
\usepackage{mathtools}
\usepackage{amsthm}

\usepackage[capitalize,noabbrev]{cleveref}

\theoremstyle{plain}

\theoremstyle{definition}

\theoremstyle{remark}

\usepackage[textsize=tiny]{todonotes}

\usepackage{hyperref}
\usepackage{url}
\usepackage{booktabs}
\usepackage{amsmath}
\usepackage{graphicx} 
\usepackage{multirow}
\usepackage{array} 
\usepackage{tabularx} 
\usepackage{enumerate}
\usepackage{enumitem}
\usepackage{xspace}
\usepackage{pifont}
\usepackage{makecell} 

\usepackage[skip=2pt]{caption}

\newcommand{\MODEL}{CRIB\xspace}

\usepackage[capitalize]{cleveref}
\crefname{section}{Sec.}{Secs.}
\Crefname{section}{Section}{Sections}
\crefname{table}{Tab.}{Tabs.}
\Crefname{table}{Table}{Tables}
\crefformat{equation}{Eq.~#2#1#3}

\newcommand{\myfirst}[1]{\textcolor{red}{\textbf{#1}}}
\newcommand{\mysecond}[1]{\textcolor{blue}{\underline{#1}}}

\icmltitlerunning{Revisiting Multivariate Time Series Forecasting with Missing Values}

\begin{document}

\twocolumn[
  \icmltitle{Revisiting Multivariate Time Series Forecasting with Missing Values}



  
  \icmlsetsymbol{NUIntern}{*}

  \begin{icmlauthorlist}
    \icmlauthor{Jie Yang}{UIC,NUIntern}
    \icmlauthor{Yifan Hu}{Tsinghua}
    \icmlauthor{Kexin Zhang}{NU}
    \icmlauthor{Luyang Niu}{Tongji}
    \icmlauthor{Philip S. Yu}{UIC}
    \icmlauthor{Kaize Ding}{NU}
  \end{icmlauthorlist}


  \icmlaffiliation{UIC}{University of Illinois Chicago}
  \icmlaffiliation{Tsinghua}{Tsinghua University}
  \icmlaffiliation{Tongji}{Tongji University}
  \icmlaffiliation{NU}{Northwestern University}

  \icmlcorrespondingauthor{Kaize Ding}{kaize.ding@northwestern.edu}

  \icmlkeywords{Machine Learning, ICML}

  \vskip 0.3in
]



\printAffiliationsAndNotice{\textsuperscript{*}Work done during an internship at Northwestern University.}

\begin{abstract}
  Missing values are common in real-world time series, and multivariate time series forecasting with missing values~(MTSF-M) has become a crucial area of research for ensuring reliable predictions.
  To address the challenge of missing data, current approaches have developed an imputation-then-prediction framework that uses imputation modules to fill in missing values, followed by forecasting on the imputed data.
  However, this framework overlooks a critical issue: there is no ground truth for the missing values, making the imputation process susceptible to errors that can degrade prediction accuracy.
  In this paper, we conduct a systematic empirical study and reveal that imputation without direct supervision can corrupt the underlying data distribution and actively degrade prediction accuracy. 
  To address this, we propose a paradigm shift that moves away from imputation and directly predicts based on the partially observed time series. 
  We introduce \textbf{C}onsistency-\textbf{R}egularized \textbf{I}nformation \textbf{B}ottleneck~(\MODEL), a novel framework built on the Information Bottleneck principle. 
  \MODEL combines a unified-variate attention mechanism with a consistency regularization scheme to learn robust representations that filter out noise introduced by missing values while preserving essential predictive signals.
  Comprehensive experiments on several real-world datasets demonstrate the effectiveness of \MODEL, which predicts accurately even under high missing rates.
  Our code is available in \href{https://github.com/Muyiiiii/CRIB}{https://github.com/Muyiiiii/CRIB}.
\end{abstract}

\section{Introduction}

Multivariate time series forecasting~(MTSF), which aims to predict future values of multiple variates based on historical observations, plays an important role in many domains, such as traffic flow forecasting~\citep{shang2022new, yu2017spatio, bai2020adaptive}, financial analysis~\citep{schaffer2021interrupted, zhu2024lsr, li2025r, finmamba, fintsb}, and weather prediction~\citep{zheng2015forecasting, wu2021autoformer, tan2022new}.
However, due to uncontrollable factors such as data collection difficulties and transmission failures~\citep{li2023missing, marisca2022learning, cini2021filling, zhang2025cross}, real-world multivariate time series data is often partially observed, with missing values scattered throughout the series.
These missing values inevitably introduce noise, 
leading to distribution shifts and disrupting the variate correlations.
MTSF models~\citep{cao2020spectral, liu2022pyraformer, ekambaram2023tsmixer, timefilter}, which typically rely on complete data, are highly sensitive to such shifts and correlation destruction, thus failing to make accurate predictions~\citep{zhou2023sloth, amd}.
This has driven increasing interest in multivariate time series forecasting with missing values~(MTSF-M)~\citep{cao2018brits, zuo2023graph, tang2020joint}, where the objective is to generate accurate and robust forecasts despite the presence of incomplete data.

To mitigate the impact of missing values, recent MTSF-M research~\citep{yu2025ginar+, peng2025s4m} has focused on enhancing observed data by imputing missing values to improve prediction performance.
One common approach is the two-stage framework,
where an imputation module~\citep{wu2022timesnet, cao2018brits, du2023saits} first fills in the missing values, 
and a forecasting model then predicts future values based on the imputed data~\citep{peng2025s4m, chen2023biased, wu2015time}.
Moreover, to reduce error accumulation between these two stages of two separate models, some studies have proposed an end-to-end framework~\citep{yu2024ginar, yu2025ginar+} that imputes missing values progressively during encoding and performs forecasting using the imputed representations. 
Overall, these methods generally follow an imputation-then-prediction paradigm, aiming to improve forecasting accuracy by mitigating the negative effects of missing values compared to directly applying forecasting models to incomplete data.

\begin{figure*}[t]
  \centering
  \includegraphics[width=0.95\textwidth]{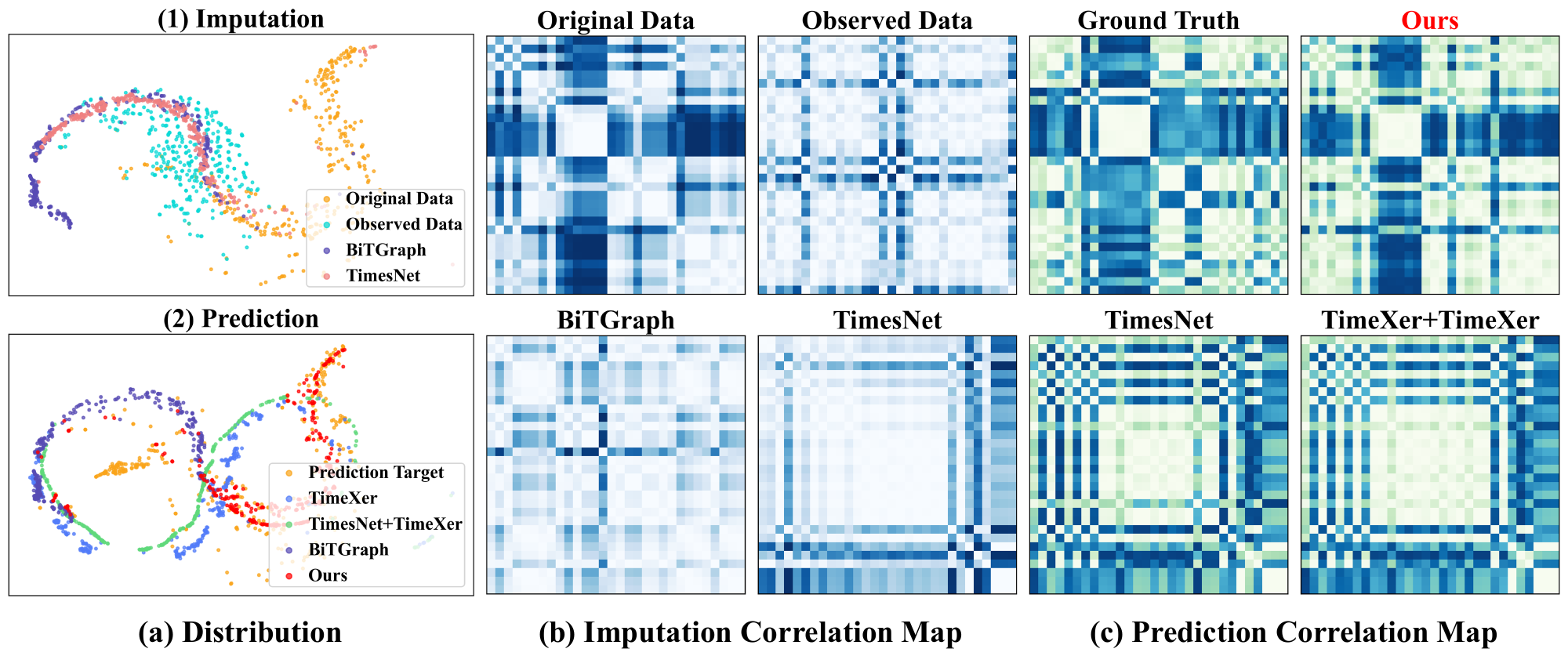}
  \caption{
      Analysis of the imputation-then-prediction paradigm on PEMS-BAY (40\% missing rate).
      \textbf{(a)} t-SNE visualizations show that current imputation modules cannot recover the original data distribution and their forecasts mismatch with the prediction target, while our direct-prediction method aligns better with the target. 
      \textbf{(b, c)} Correlation maps reveal that imputation fails to recover true variate correlations, whereas our method preserves underlying correlations more effectively. 
      Quantitative analysis is in App.~\ref{app:Quantitative_Analysis_of_Introduction}.
  }
  \label{fig:intro_analysis}
\end{figure*}

However, current MTSF-M methods ignore a critical limitation in real-world applications: \textbf{there is no ground truth for missing values}.
In such scenarios, the imputation module of the current MTSF-M methods would lack reliable guidance, which means the imputed values and reconstructed correlations cannot be guaranteed to be accurate with only the final prediction guidance.
As a result, noise would propagate into the prediction stage and degrade forecasting performance, particularly when the missing rate is high. 
To investigate this issue, we conduct an empirical analysis of representative imputation-then-prediction methods, where original and observed data denote the complete and partially observed data, respectively.
This includes the two-stage framework combining TimesNet~\citep{wu2022timesnet} for imputation and TimeXer~\citep{wang2024timexer} for forecasting, as well as the end-to-end framework BiTGraph~\citep{chen2023biased}.
\cref{fig:intro_analysis} illustrates the empirical results, where panel~(a) visualizes the distributions of imputed and predicted values, and panels~(b) and~(c) present the correlations among variates.  
Our findings highlight two key phenomena:

\begin{itemize}[leftmargin=*]
    \item[\ding{182}] \textit{\textbf{Improper imputation can corrupt the observed data.}}
    Current MTSF-M frameworks commonly employ imputation modules to recover missing values. 
    However, as shown in \cref{fig:intro_analysis}~(a-1, b), without enough direct supervision, imputed values deviate considerably from the distribution of the original complete data, and the underlying correlations among variates are not correctly reconstructed.
    The deterioration of both the data distribution and variate correlations suggests that imputation guided only by the prediction objective can degrade the observed data rather than effectively repairing the missing values.
    
    \item[\ding{183}] \textit{\textbf{Flawed imputation, in turn, leads to poor prediction performance.}} 
    Errors from the imputation stage inevitably propagate into forecasting. 
    As shown in \cref{fig:intro_analysis}~(a-2, c), the predictions exhibit large deviations from the prediction targets.
    Notably, even a model TimeXer applied directly to incomplete observed data outperforms a more complex framework that combines TimesNet for imputation with TimeXer for prediction. 
    These findings indicate that a flawed imputation stage can actively harm, rather than enhance, the forecasting capabilities of a model.
\end{itemize}

Based on these two observations, we ask a fundamental question: \textit{\textbf{Is it possible to predict directly from partially observed time series, avoiding the pitfalls of imputation while maintaining high accuracy?}} 
To answer this, we propose \textbf{C}onsistency-\textbf{R}egularized \textbf{I}nformation \textbf{B}ottleneck~(\MODEL),
a novel framework that predicts directly from partially observed data, bypassing the issues associated with imputation.
CRIB is built on the Information Bottleneck~(IB) principle, which enables it to learn a compressed representation that filters noise from missing values while preserving essential predictive signals.
To achieve this, it employs a unified-variate attention mechanism to capture complex correlations from the sparse input and is trained with a consistency regularization scheme to enhance robustness, especially under high missing rates.

Our main contributions can be summarized as follows: 
\begin{itemize}[leftmargin=*]

\item \textbf{Empirical analysis}: 
We perform a systematic empirical study of dominant imputation-then-prediction paradigms for MTSF-M,
showing that imputation modules guided only by the prediction objective can corrupt the observed data distribution and degrade prediction performance.

\item \textbf{Method}: 
We propose a novel direct-prediction method, \MODEL, which removes the imputation step completely.
Built on the Information Bottleneck principle, \MODEL integrates unified-variate attention and consistency
regularization to learn refined representations that balance noise filtering and task-relevant information preservation.

\item \textbf{Experiments}: 
Extensive experiments on four real-world benchmarks demonstrate that \MODEL significantly outperforms
state-of-the-art methods by an average of 18\%, particularly under high missing rates,
highlighting the advantage of direct prediction over imputation-then-prediction.

\end{itemize}

\section{Preliminaries}

\paragraph{Notations \& Problem Formulation} 
In MTSF-M tasks, the historical time series is denoted as $X = \{ x_i^{1:T}\ |\ i=1,\cdots, N \}\in \mathbb{R}^{N\times T}$, where $T$ is the number of time steps and $N$ is the number of variates. The goal is to predict the future $S$ time steps $Y =\{ x_{i}^{T+1:T+S}\ |\ i=1, \cdots, N \}\in \mathbb{R}^{N\times S}$.
Missingness is represented by a binary mask $M\in \{0,1\}^{N\times T}$, where $X^{\text{o}}=\{ X^{i,j}|M^{i,j}=1\}$ are observed values and $X^\text{m}=\{ X^{i,j}|M^{i,j}=0\}$ are missing values.
We denote $Z\in \mathbb{R}^{N\times D}$ as the intermediate representations of input, where $D$ is the dimension of the representation.

\paragraph{Information Bottleneck for MTSF-M}
IB theory~\citep{tishby2015deep, voloshynovskiy2019information, yang2025glocal} provides an information-theoretic framework for learning compact yet informative representations.
Given the partially observed input $X^{\text{o}}$ and prediction target $Y$, the goal is to learn a latent representation $Z$ that compresses $X^{\text{o}}$
while preserving maximal information about $Y$.
This trade-off in \MODEL is formalized as follows:
\begin{equation}
    \min_{\theta}\ [I_{\theta}(Z;X^{\text{o}})-\beta \cdot I_{\theta}(Y;Z)].
    \label{eq:IB-loss}
\end{equation}
Here, $\theta$ represents the learnable parameters of our proposed \MODEL.
$I(Z;X^{\text{o}})$ and $I(Y;Z)$ are the mutual information terms measuring compactness and informativeness, respectively. 
The Lagrange multiplier $\beta \in \mathbb{R}^{+} $ controls the balance between these two terms~\citep{tishby2000information}.
Furthermore, under standard assumptions in the IB literature~\citep{alemi2016deep, chalk2016relevant, ma2023temporal}, the joint distribution of the variables can be factorized as:
\begin{equation}
\begin{aligned}
p(X^{\text{o}},Y,Z)
&= p(Z \mid X^{\text{o}},Y)\, p(Y \mid X^{\text{o}})\, p(X^{\text{o}}) \\
&= p(Z \mid X^{\text{o}})\, p(Y \mid X^{\text{o}})\, p(X^{\text{o}}),
\end{aligned}
\label{eq:XYZRelation}
\end{equation}
namely, there is a Markov chain $Y\leftrightarrow X^{\text{o}} \leftrightarrow Z$, indicating that the representations $Z$ is learned only from $X^{\text{o}}$ without direct access to the target $Y$. 

\section{Methodology}

\begin{figure*}[htbp]
    \centering
    \includegraphics[width=0.95\textwidth]{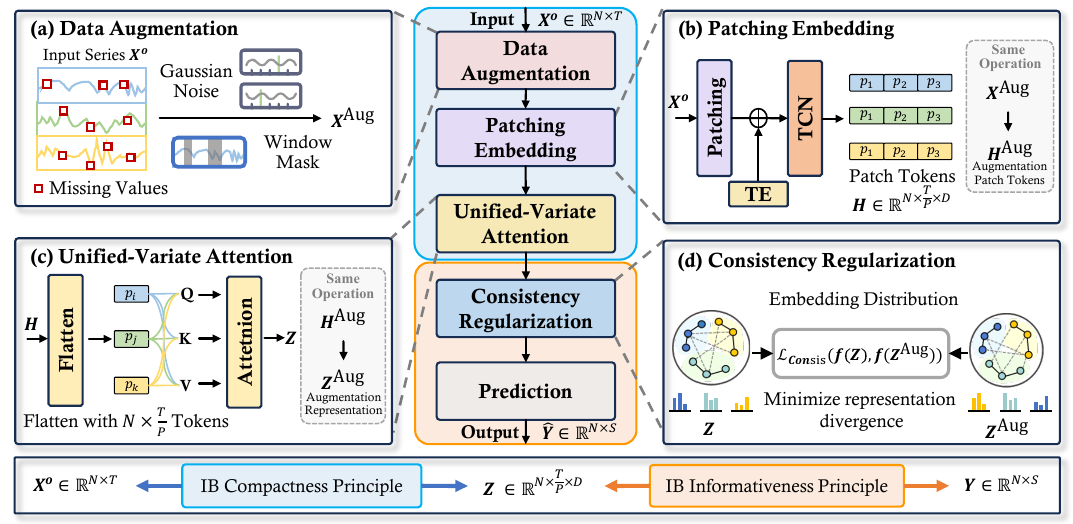}
    \caption{
    Overall framework of \MODEL.
    \textbf{(a)} Data Augmentation creates a more challenging view of the partially observed data $X^{\text{o}}$ by generating an augmented version $X^{\text{Aug}}$. 
    \textbf{(b)} The Patching Embedding layer converts the $X^{\text{o}}$ and $X^{\text{Aug}}$ into robust patch-level feature representations $H$ and $H^{\text{Aug}}$. 
    \textbf{(c)} The Unified-Variate Attention mechanism models the global correlations between all the patches within $H$ and $H^{\text{Aug}}$ to produce refined representations $Z$ and $Z^{\text{Aug}}$. 
    \textbf{(d)} Consistency Regularization aligns the representations from the original $Z$ and the augmented views $Z^{\text{Aug}}$.
    The entire process is guided by the IB principles of compactness and informativeness to predict~$\widehat{Y}$.
    }
    \label{fig:Model}
\end{figure*}

As illustrated in \cref{fig:Model}, our proposed model, \MODEL, bypasses the problematic imputation stage by performing forecasts directly on the partially observed data.
The architecture is composed of several key stages, each designed to address the challenges of learning from partially observed data.
First, to handle the raw, sparse input, we introduce a Patching Embedding layer that employs a Temporal Convolutional Network~(TCN)~\citep{bai2018empirical} to learn robust local feature representations from available data points.
Second, to capture the complex global correlations that are disrupted by missingness, a Unified-Variate Attention mechanism models correlations across all patches simultaneously.
Third, to ensure the model learns features that are stable and invariant to different missingness, especially under high missing rates, we introduce a Consistency Regularization scheme based on data augmentation.
The entire learning process is guided by the IB principle, which provides a theoretical foundation for learning a representation that is maximally compressive against noise while being sufficiently informative for the forecasting task.


\subsection{Patching Embedding}


To effectively enhance the semantic information that is not available in the partially observed, point-level time series $X^{\text{o}}\in \mathbb{R}^{N\times T}$, we first transform the input into a sequence of more meaningful patch-level representations~\citep{nie2022time}.
The series is partitioned into non-overlapping patches $\widehat{X}=\{\widehat{x}_{i}^{1:T/P}\ |\ i=1,\cdots, N  \}\in \mathbb{R}^{N\times (T/P)\times P}$ of length $P$.
We choose $P$ such that the total length $T$ is evenly divisible. 
Consequently, this patching strategy reduces the sequence length from $T$ to $T/P$, thus remarkably lowering the memory and computational cost of attention calculation.

Next, to enable the following unified-variate attention mechanism to capture the temporal directionality of each variate
$x_i^{1:T}$, we adopt the temporal encoding strategy inspired by vanilla transformer~\citep{vaswani2017attention} as follows:
\begin{equation}
    \text{TE}(t,m)=
    \begin{cases}
      \ \sin(t/10000^{2m/P}) & \text{ if } m=2k, \\
      \ \cos(t/10000^{2m/P}) & \text{ if } m=2k+1,
    \end{cases}
\end{equation}
where $m$ represents the $m$-th dimension of the feature.
These temporal embeddings are added to the input patches to provide temporal information.
Each patch, now containing a mix of observed values and temporal embeddings, is then processed by a TCN. 
It utilizes its efficient dilated convolution structure to transform sparse patches with missing values into dense feature representations $H\in \mathbb{R}^{N\times (T/P)\times D}$ that capture local temporal correlations. 


\subsection{Unified-Variate Attention}

To model the complex, non-local correlations disrupted by missing data, we introduce a unified attention mechanism. 
Instead of using separate modules for inter- and intra-variate correlations among all the variates, our approach treats all patch representations uniformly. 
We first flatten the patch representations $H$ into a sequence $\widehat{H}\in \mathbb{R}^{(N\times T/P)\times D}$ with $N\times T/P$ tokens. 
A standard self-attention mechanism is then applied to this flattened sequence:
\begin{equation}
    Z = \text{Attention}(Q,K,V)=\text{Softmax}(\frac{QK^{\top}}{\sqrt{D}} )V, 
\end{equation}
where $Q, K, V\in \mathbb{R}^{(N\times T/P)\times D} $ are the linear projections of tokens $\widehat{H}$, and $\top$ denotes the matrix transpose.
This allows the model to learn all possible correlations—both within a single variate's timeline~(intra-variate) and across different variates~(inter-variate)—without imposing strong, predefined structural biases.
Such flexibility is particularly advantageous for sparse data, as it permits the model to rely on the most informative available signals, regardless of their origin.
Unlike previous methods~\citep{yi2024fouriergnn, wang2024fully} that employ strategies to reduce the memory and time costs of attention calculations, often at the expense of attention mechanism performance, we accelerate attention computation by patching time series.
This can reduce the number of temporal tokens from $T$ to $T/P$, lowering the memory and computational cost of attention calculation by a factor of $P^2$, while enhancing the semantic-level information of the data. 

\subsection{Final Prediction} 

In \MODEL, we implement the predictor using a simple Multi-Layer Perceptron~(MLP) as follows:
\begin{align}
    &\widehat{Y}=\text{Predictor}(Z)=\text{MLP}(Z)\in \mathbb{R}^{N\times S},
\end{align}
where $S$ is the prediction length and $\text{MLP}(\cdot)$ denotes a simple two-layer fully connected network with a ReLU activation function applied between the layers.
We deliberately employ a simple linear predictor to demonstrate that the forecasting performance of CRIB stems from the high-quality, robust representations $Z$ learned by our IB-guided attention mechanism, rather than employing a complex and powerful predictor~\citep{liu2023itransformer, zeng2023transformers}.



\subsection{Information Bottleneck Guidance}

To enhance the quality of the learned representations $Z$ and improve forecasting accuracy, we propose an IB-based guidance. 
This guidance aims to balance compactness (filtering out irrelevant information) with informativeness (preserving relevant task-specific signals), allowing \MODEL to focus on the most significant factors for accurate forecasting.
In this section, we present how the compactness and informativeness principles are formulated and implemented in our framework.
Full derivations are detailed in App.~\ref{app:FullDerivation}.

\subsubsection{Compactness Principle}
The compactness principle, which aims to minimize the mutual information $I_{\theta}(Z;X^{\text{o}})$, forces the learned representation $Z$ to be a minimal sufficient statistic of the input. 
In our context, this encourages the model to discard non-essential information, which critically includes the noise introduced by the arbitrary locations of missing values.
Following the variational inference~\citep{voloshynovskiy2019information}, we derive a equivalent form of the compactness term in \cref{eq:IB-loss}:
\begin{equation}
\begin{aligned}
I_{\theta}(Z;X^{\text{o}})
&= \mathbb{E}_{p(x^{\text{o}},z)}
\!\left[
\log \frac{p(x^{\text{o}},z)}{p(z)\,p(x^{\text{o}})}
\right] \\
&= \mathbb{E}_{p(x^{\text{o}})}
\!\left[
D_{\mathrm{KL}}\!\left(p(z \mid x^{\text{o}})\,\|\,q(z)\right)
\right] \\
&\quad - D_{\mathrm{KL}}\!\left(p(z)\,\|\,q(z)\right).
\end{aligned}
\label{eq:IZX1}
\end{equation}
Because of difficulty in posterior calculation and the non-negative property of Kullback-Leibler~(KL) divergence, we use $p_{\theta}(z| x^{\text{o}})$ to approximate the true posterior distribution $p(z|x^{\text{o}})$ and bound \cref{eq:IZX1}:
\begin{equation}
    \begin{aligned}
        I_{\theta}(Z;X^{\text{o}})
        &\le \mathbb{E}_{p(x^{\text{o}})}D_{KL}[p_{\theta}(z|x^{\text{o}})||q(z)]
        \stackrel{\text{def}}{=} \mathcal{L}_{\text{Comp}},
    \end{aligned}
    \label{eq:IZX2}
\end{equation}
where we set isotropic Gaussian as the prior distribution of refined representations $Z$, i.e., $p(Z)=\mathcal{N}(0, I)$.
Therefore, representations~$Z$ are produced through a multivariate Gaussian distribution as:
\begin{equation}
    p_{\theta}(Z|X^{\text{o}})=\mathcal{N}({\mu}_{\theta}(X^{\text{o}}),\text{diag}(\delta_{\theta}(X^{\text{o}}))),
\end{equation}
where $\mu_{\theta}(\cdot)$ and $\sigma_{\theta}(\cdot)$ are designed as neural networks with parameter $\theta$.
For training, we use the standard reparameterization trick~\citep{kingma2013auto}, $Z = \mu_{\theta}(X^{\text{o}}) + \sigma_{\theta}(X^{\text{o}}) \odot \epsilon$, which makes the objective in \cref{eq:IZX2} differentiable without the need for stochastic estimation as follows:
\begin{equation}
\begin{aligned}
\mathcal{L}_{\text{Comp}}
&= \frac{1}{2} \sum_{j=1}^{D} \Big(
    1 + \log \big(\sigma_{\theta}^{(j)}(X^{\text{o}})\big)^{2} \\
&\qquad
    - \big(\mu_{\theta}^{(j)}(X^{\text{o}})\big)^{2}
    - \big(\sigma_{\theta}^{(j)}(X^{\text{o}})\big)^{2}
\Big).
\end{aligned}
\end{equation}
Here, $\mu_{\theta}^{(j)}(X^{\text{o}})$ and $\sigma_{\theta}^{(j)}(X^{\text{o}})$ denote the $j$-th element of the mean and standard deviation vectors.

\subsubsection{Informativeness Principle}

To balance the compactness objective, the informativeness principle ensures that the representation $Z$ preserves sufficient information for the forecasting task. 
To derive a tractable lower bound for the informativeness term, we follow the framework in~\citep{voloshynovskiy1912information} and \cref{eq:XYZRelation}, and assume that prediction errors follow a Gaussian distribution with fixed variance~$\sigma^2$, i.e., $q_{\theta}(y|z)=\mathcal{N}(\widehat{y},\sigma^2I)$~\citep{choi2023conditional}.
The derivation proceeds as:
\begin{equation}
\begin{aligned}
I_{\theta}(Y;Z)
&= \mathbb{E}_{p(z,y)}
\left[\log \frac{q_{\theta}(y|z)}{p(y)}\right]
+ \mathbb{E}_{p(z,y)}
\left[\log \frac{p(y|z)}{q_{\theta}(y|z)}\right] \\
&\ge \mathbb{E}_{p(z,y)}[\log q_{\theta}(y|z)] \\
&= - \mathbb{E}_{p(z,y)}
\left[
    \frac{1}{2\sigma^2} \|y - \widehat{y}\|^2
    + \frac{T}{2} \log(2\pi \sigma^2)
\right] \\
&\propto - \mathbb{E}_{p(z,y)} \left[\|y - \widehat{y}\|^2\right]
\stackrel{\text{def}}{=} -\mathcal{L}_{\text{Pred}} .
\end{aligned}
\end{equation}
thus encouraging the model to extract task-relevant information from intermediate representations.

\subsection{Consistency Regularization}

While the IB framework encourages learning a compact representation, high missing rates can still lead to unstable training as shown in App.~\ref{app:IB-Attention}, where the model overfits to the specific variate in a given time window~\citep{choi2023conditional}.
To mitigate this and enhance robustness, we introduce a consistency regularization scheme~\citep{bachman2014learning, laine2016temporal}.
The core intuition is that the model's prediction should be invariant to the missingness.
We achieve this by creating an augmented, more challenging view of the input, e.g, introducing additional noise to partially observed data.
By enforcing that the representations learned from the observed and augmented views remain consistent, we regularize the model to handle missing values while stabilizing the refined representations instead of focusing excessively on a limited subset of observed data and neglecting crucial task-relevant variate correlations.

\paragraph{Data Augmentation}
Specifically, we generate $X^{\text{Aug}}\in\mathbb{R}^{N\times T}$ by applying two augmentations~\citep{wen2020time}: 
(1) Random Masking, where we randomly select an additional 10\% of the observed time points and set them to zero to simulate a more severe missingness scenario; 
and (2) Gaussian Noise, where we add noise $\epsilon \in \mathcal{N}(0,I)$ to all observed points to simulate sensor noise, enhancing the model's robustness to minor fluctuations in the input..

\paragraph{Consistency Regularization}
Then, through the same forward process as $X^{\text{o}}$, we can get their refined representations $Z^{\text{Aug}}$. 
The refined representations of observed and augmented data are regularized via the following consistency regularization loss function:
\begin{equation}
    \mathcal{L}_{\text{Consis}}=\frac{1}{N\times T/P} \sum_{i=1}^{N\times T/P}||z_i-z^{\text{Aug}}_i||^2,
\end{equation}
where $N\times T/P$ is the number of the flattened tokens.
By aligning the representations of the observed and augmented data, the model is encouraged to learn stable representations, thus enhancing robustness in scenarios with high missing rates. 
Furthermore, this consistency regularization can be seamlessly integrated into the overall optimization objective, complementing the IB theory to ensure that the refined representations retain essential task-relevant information while filtering out irrelevant noise from the missing values.

\subsection{Model Learning}

\renewcommand{\arraystretch}{1.25}
\begin{table*}[!htbp]
\centering
\caption{Performance comparison on four datasets with a point missing pattern~(average MAE and MSE across 20\% to 70\% missing rate). 
Best is \myfirst{Bold} and second-best is \mysecond{Underlined}.
}
\label{tab:MainExp_avg_Transposed}
\begin{small}
\renewcommand{\multirowsetup}{\centering}
\setlength{\tabcolsep}{1pt}
\resizebox{\textwidth}{!}{
\begin{tabular}{lc|c|c|c|c|c|cc|cc|cc|cc|cc|cc|cc|c|c}
\toprule
\multirow{2}{*}{\textbf{\scalebox{0.83}{Data}}} & \multirow{2}{*}{\textbf{\scalebox{0.80}{Metric}}} & \multicolumn{1}{c}{\scalebox{0.81}{BiTGraph}} & \multicolumn{1}{c}{\scalebox{0.81}{BRITS}} & \multicolumn{1}{c}{\scalebox{0.81}{GRIN}} & \multicolumn{1}{c}{\scalebox{0.81}{SAITS}} & \multicolumn{1}{c}{\scalebox{0.81}{SPIN}} & \multicolumn{2}{c}{\scalebox{0.81}{SegRNN}} & \multicolumn{2}{c}{\scalebox{0.81}{WPMixer}} & \multicolumn{2}{c}{\scalebox{0.81}{iTransformer}} & \multicolumn{2}{c}{\scalebox{0.81}{PatchTST}} & \multicolumn{2}{c}{\scalebox{0.81}{DLinear}} & \multicolumn{2}{c}{\scalebox{0.81}{TimeXer}} & \multicolumn{2}{c}{\scalebox{0.81}{PAttn}} & \scalebox{0.9}{\textbf{Ours}} & \multirow{2}{*}{\textbf{\scalebox{0.83}{IMP}}} \\

\cmidrule(lr){3-3} \cmidrule(lr){4-4} \cmidrule(lr){5-5} \cmidrule(lr){6-6} \cmidrule(lr){7-7} \cmidrule(lr){8-9} \cmidrule(lr){10-11} \cmidrule(lr){12-13} \cmidrule(lr){14-15} \cmidrule(lr){16-17} \cmidrule(lr){18-19} \cmidrule(lr){20-21} \cmidrule(lr){22-22}

& & \scalebox{0.81}{Original} & \scalebox{0.81}{Original} & \scalebox{0.81}{Original} & \scalebox{0.81}{Original} & \scalebox{0.81}{Original} & \scalebox{0.81}{Original} & \scalebox{0.81}{Imputed} & \scalebox{0.81}{Original} & \scalebox{0.81}{Imputed} & \scalebox{0.81}{Original} & \scalebox{0.81}{Imputed} & \scalebox{0.81}{Original} & \scalebox{0.81}{Imputed} & \scalebox{0.81}{Original} & \scalebox{0.81}{Imputed} & \scalebox{0.81}{Original} & \scalebox{0.81}{Imputed} & \scalebox{0.81}{Original} & \scalebox{0.81}{Imputed} & \scalebox{0.81}{Original} & \\

\midrule

\multirow{2}{*}{\scalebox{0.9}{\textbf{PEMS-BAY}}}
& \scalebox{0.80}{MAE} & 0.413 & 0.366 & 0.350 & OOM & 0.402 & 0.120 & 0.178 & 0.155 & 0.201 & \mysecond{0.107} & 0.125 & 0.129 & 0.139 & 0.156 & 0.148 & 0.125 & 0.135 & 0.110 & 0.148 & \myfirst{0.093} & \textbf{13\%} \\
& \scalebox{0.80}{MSE} & 0.788 & 0.705 & 0.623 & OOM & 0.649 & 0.067 & 0.203 & 0.082 & 0.140 & 0.055 & 0.072 & 0.060 & 0.086 & 0.087 & 0.081 & \mysecond{0.051} & 0.073 & 0.061 & 0.091 & \myfirst{0.043} & \textbf{15\%} \\

\midrule

\multirow{2}{*}{\scalebox{0.9}{\textbf{Metr-LA}}}
& \scalebox{0.80}{MAE} & 0.445 & 0.366 & 0.389 & 0.451 & 0.625 & 0.318 & 0.314 & 0.356 & 0.342 & \mysecond{0.273} & 0.290 & 0.313 & 0.306 & 0.399 & 0.366 & 0.321 & 0.298 & 0.302 & 0.294 & \myfirst{0.262} & \textbf{4\%} \\
& \scalebox{0.80}{MSE} & 0.760 & 0.611 & 0.653 & 0.721 & 0.965 & 0.345 & 0.360 & 0.356 & 0.385 & 0.317 & 0.330 & 0.320 & 0.349 & 0.373 & 0.362 & \mysecond{0.313} & 0.333 & 0.337 & 0.345 & \myfirst{0.301} & \textbf{4\%} \\

\midrule

\multirow{2}{*}{\scalebox{0.9}{\textbf{ETTh1}}}
& \scalebox{0.80}{MAE} & 0.337 & 0.357 & 0.356 & 0.372 & 0.437 & 0.356 & 0.425 & 0.340 & 0.399 & 0.342 & 0.419 & 0.324 & 0.386 & 0.402 & 0.598 & \mysecond{0.314} & 0.347 & 0.341 & 0.432 & \myfirst{0.256} & \textbf{18\%} \\
& \scalebox{0.80}{MSE} & 0.387 & 0.421 & 0.400 & 0.457 & 0.468 & 0.479 & 0.477 & 0.432 & 0.417 & 0.408 & 0.473 & 0.385 & 0.435 & 0.560 & 0.682 & 0.377 & \mysecond{0.370} & 0.416 & 0.470 & \myfirst{0.269} & \textbf{27\%} \\

\midrule

\multirow{2}{*}{\scalebox{0.9}{\textbf{Electricity}}}
& \scalebox{0.80}{MAE} & 0.036 & 0.035 & 0.034 & 0.053 & 0.136 & 0.078 & 0.255 & 0.049 & 0.218 & 0.034 & 0.130 & 0.036 & 0.105 & 0.074 & 0.210 & \mysecond{0.029} & 0.083 & 0.042 & 0.152 & \myfirst{0.026} & \textbf{10\%} \\
& \scalebox{0.80}{MSE} & 0.113 & 0.059 & 0.061 & 0.266 & 0.358 & 1.010 & 1.286 & 0.172 & 0.286 & \mysecond{0.054} & 0.547 & 0.092 & 0.379 & 0.404 & 2.000 & 0.064 & 0.100 & 0.115 & 0.864 & \myfirst{0.044} & \textbf{18\%} \\
\bottomrule
\end{tabular}
}
\end{small}
\end{table*}

\renewcommand{\arraystretch}{1.25}
\begin{table}[!htbp]
\centering
\caption{Performance comparison on six additional datasets with a point missing pattern~(average MAE and MSE across 20\% to 70\% missing rate).
Best is \myfirst{Bold} and second-best is \mysecond{Underlined}.
}
\label{tab:MainExp_avg_Transposed_new6}
\begin{small}
\renewcommand{\multirowsetup}{\centering}
\setlength{\tabcolsep}{1pt}
\resizebox{\linewidth}{!}{
\newcolumntype{C}{>{\centering\arraybackslash}p{1.1cm}}
\begin{tabular}{lc|C|C|C|C|C|C|C|C|C|C|C}
\toprule
\textbf{Data} & \textbf{Metric} & \scalebox{0.75}{SegRNN} & \scalebox{0.75}{WPMixer} & \scalebox{0.65}{iTransformer} & \scalebox{0.75}{PatchTST} & \scalebox{0.75}{DLinear} & \scalebox{0.75}{PAttn} & \scalebox{0.75}{CSDI} & \scalebox{0.65}{NeuralCDE} & \scalebox{0.6}{ImputeFormer} & \scalebox{0.75}{TimesNet} & \scalebox{0.75}{\textbf{Ours}} \\
\midrule
\multirow{2}{*}{\textbf{ETTh2}}
& MAE & 0.163 & 0.171 & 0.225 & 0.182 & 0.225 & 0.211 & 0.519 & 0.320 & \mysecond{0.152} & 0.182 & \myfirst{0.141} \\
& MSE & 0.061 & 0.066 & 0.124 & 0.090 & 0.117 & 0.106 & 0.684 & 0.188 & \mysecond{0.054} & 0.076 & \myfirst{0.048} \\
\midrule
\multirow{2}{*}{\textbf{ETTm1}}
& MAE & \mysecond{0.339} & 0.381 & 0.398 & 0.377 & 0.505 & 0.389 & 0.658 & 0.498 & 0.400 & 0.361 & \myfirst{0.313} \\
& MSE & \mysecond{0.526} & 0.637 & 0.671 & 0.606 & 0.967 & 0.638 & 1.284 & 0.853 & 0.601 & 0.565 & \myfirst{0.453} \\
\midrule
\multirow{2}{*}{\textbf{ETTm2}}
& MAE & 0.130 & 0.153 & 0.206 & 0.182 & 0.217 & 0.190 & 0.527 & 0.234 & \mysecond{0.123} & 0.170 & \myfirst{0.116} \\
& MSE & 0.041 & 0.054 & 0.113 & 0.092 & 0.107 & 0.098 & 0.681 & 0.110 & \mysecond{0.037} & 0.065 & \myfirst{0.034} \\
\midrule
\multirow{2}{*}{\textbf{Weather}}
& MAE & \mysecond{0.059} & 0.087 & 0.118 & 0.121 & 0.171 & 0.118 & 0.381 & 0.131 & \mysecond{0.059} & 0.111 & \myfirst{0.055} \\
& MSE & \mysecond{0.052} & 0.052 & 0.140 & 0.120 & 0.155 & 0.114 & 0.882 & 0.133 & \mysecond{0.052} & 0.081 & \myfirst{0.048} \\
\midrule
\multirow{2}{*}{\textbf{Exchange}}
& MAE & 0.088 & 0.108 & 0.187 & 0.125 & 0.216 & 0.091 & 0.626 & 0.401 & \mysecond{0.056} & 0.115 & \myfirst{0.034} \\
& MSE & 0.025 & 0.031 & 0.082 & 0.065 & 0.124 & 0.052 & 1.109 & 0.337 & \mysecond{0.007} & 0.036 & \myfirst{0.003} \\
\midrule
\multirow{2}{*}{\textbf{BeijingAir}}
& MAE & \mysecond{0.358} & 0.371 & 0.382 & 0.368 & 0.405 & 0.381 & 0.543 & 0.441 & 0.366 & 0.376 & \myfirst{0.348} \\
& MSE & \myfirst{0.532} & 0.587 & 0.598 & 0.568 & 0.619 & 0.592 & 1.018 & 0.691 & 0.564 & 0.585 & \mysecond{0.533} \\
\bottomrule
\end{tabular}
}
\end{small}
\end{table}

We have proposed a consistency-regularized method \MODEL, which can complete MTSF-M tasks based on the IB theory.
Overall, we optimize our model based on the following objective by combining all the introduced loss functions:
\begin{equation}
    \min_{\theta}\ [\alpha\cdot(\mathcal{L}_{\text{Comp}}^{\theta}+\beta\cdot\mathcal{L}_{\text{Pred}}^{\theta})+\gamma\cdot\mathcal{L}_{\text{Consis}}],
\end{equation}
where $\alpha, \beta, \gamma \in \mathbb{R}$ are the preset balancing coefficients.
This entire guidance helps \MODEL extract the most important task-relevant information from the partially observed data while filtering out noise introduced by missing values.

\section{Experiment}

\subsection{Experiment Settings}

\paragraph{Datasets.}
We evaluate our model on 11 MTSF datasets: PEMS-BAY, Metr-LA~\citep{li2017diffusion},
ETTh1, ETTh2, ETTm1, ETTm2~\citep{zhou2021informer}, Weather~\cite{miao2024unified},
BeijingAir, Exchange~\cite{lai2018modeling}, Electricity~\citep{wu2021autoformer}, and AQI~\cite{yi2016st}.
More information about these datasets is summarized in~App.~\ref{app:Datasets}.

\paragraph{Baselines.}
We evaluate 16 representative baselines:
(1) MTSF-M methods: BRITS~\citep{cao2018brits}, SAITS~\citep{du2023saits}, SPIN~\citep{marisca2022learning}, GRIN~\citep{cini2021filling}, and BiTGraph~\citep{chen2023biased}. 
(2) Imputation methods: CSDI~\cite{tashiro2021csdi}, NeuralCDE~\cite{kidger2020neural}, ImputeFormer~\cite{nie2024imputeformer}, and TimesNet~\cite{wu2022timesnet}.
(3) Transformer-based MTSF methods: iTransformer~\citep{liu2023itransformer}, PatchTST~\citep{nie2022time}, and PAttn~\citep{tan2024language}. 
(4) MLP- and RNN-based MTSF methods: DLinear~\citep{zeng2023transformers}, WPMixer~\citep{murad2025wpmixer}, TimeXer~\citep{wang2024timexer}, and SegRNN~\citep{lin2023segrnn}.


\paragraph{Implementation Details.}
\label{subsubsec:Implementation_Details}

Several forecasting baselines are not originally designed for MTSF-M settings.
We therefore construct two-stage variants by combining them with SOTA imputation model \textbf{TimesNet}~\citep{wu2022timesnet},
which which imputes missing values before forecasting.
To simulate a practical scenario where ground truth for missing values is unavailable during inference, TimesNet is trained with a 10\% missing rate and then used to impute datasets with 20\%, 40\%, 60\%, and 70\% missing rates. 
The original models and the variants are denoted as \textbf{Original} and \textbf{Imputed}, respectively.
More details are provided in~App.~\ref{app:Comparison_Evaluations}.

\textbf{Training.} 
We evaluate forecasting performance using Mean Squared Error (MSE) and Mean Absolute Error (MAE).
Our training cost is in App.~\ref{app:Training_Cost}.

\subsection{Main Results}

\begin{figure*}[htbp]
    \centering
    \includegraphics[width=1\textwidth]{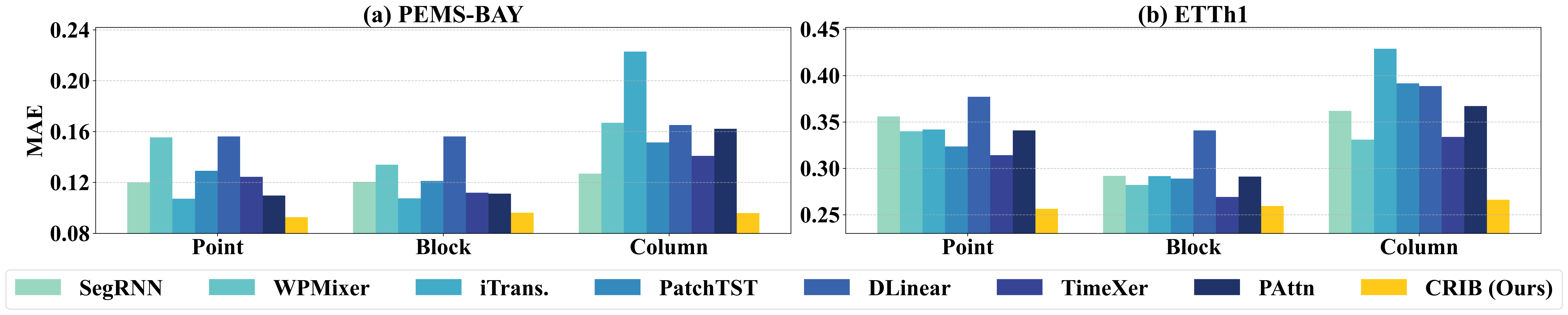}
    \caption{ Average MAE on PEMS-BAY and ETTh1 with point, block, and column missing
    patterns. }
    \label{fig:MissingPattern}
\end{figure*}

The average performance comparisons are reported in
~\cref{tab:MainExp_avg_Transposed,tab:MainExp_avg_Transposed_new6}, with full results
and additional missing-pattern evaluations provided in
App.~\ref{app:Full_Exp} and~\cref{fig:MissingPattern}.
We denote out-of-memory and
improvement as OOM and IMP, respectively. 
Based on these results, we summarize
our observations~(\textbf{Obs.}) as follows:

\textbf{Obs. \ding{182}: \MODEL demonstrates superior performance improvement in
MTSF-M tasks.} 
Overall, \MODEL consistently improves forecasting accuracy over strong baselines
across diverse datasets and missing-value settings.
Specifically, \MODEL reduces the MAE by over 18\%
on ETTh1 and over 13\% on PEMS-BAY compared to the strongest baseline. We attribute
this improvement to our model's design, which integrates patch embedding, unified-variate
attention, and consistency regularization under the IB principle, thus enabling \MODEL
to effectively filter noise from incomplete data while preserving essential predictive
signals.

\textbf{Obs. \ding{183}: Modern MTSF models have surpassed specialized models,
and applying imputation to them is often detrimental.} Our experiments show that
recent MTSF models (e.g., PatchTST), when applied directly to partially observed
data, consistently outperform methods designed specifically for missing values (e.g.,
BiTGraph). Moreover, we find that applying an explicit imputation step to these
modern models is often harmful; their performance on partially observed data is frequently
superior to that of their two-stage variants, which use a pre-trained imputer (e.g.,
TimesNet). For example, PatchTST has an average $0.324$ MAE while its variant
has a worse average $0.386$ MAE on the ETTh1 dataset. These phenomena suggest that
imputation without direct ground-truth supervision can introduce erroneous values.
This, in turn, distorts the underlying data distribution and corrupts variate
correlations, ultimately degrading forecasting performance.

\subsection{Natural Missingness Evaluation}

\begin{table}[htbp]
    \centering
    \caption{Performance comparison on AQI datasets (Original vs. Imputed). The best results are highlighted in \myfirst{Bold}, and the second-best is highlighted in \mysecond{Underline}.}
    \label{tab:aqi_results_v2}

    \resizebox{\linewidth}{!}{%
        \newcolumntype{C}{>{\centering\arraybackslash}p{1.1cm}}
        \begin{tabular}{lc|C|C|C|C|C|C|C|C|C|C|C}
            \toprule
            \textbf{Data} & \textbf{Metric} & \scalebox{0.75}{\textbf{SegRNN}} & \scalebox{0.75}{\textbf{WPMixer}} & \scalebox{0.65}{\textbf{iTransformer}} & \scalebox{0.75}{\textbf{PatchTST}} & \scalebox{0.75}{\textbf{DLinear}} & \scalebox{0.75}{\textbf{PAttn}} & \scalebox{0.75}{\textbf{CSDI}} & \scalebox{0.65}{\textbf{NeuralCDE}} & \scalebox{0.6}{\textbf{ImputeFormer}} & \scalebox{0.75}{\textbf{TimesNet}} & \scalebox{0.75}{\textbf{Ours}} \\
            \midrule
            \multirow{2}{*}{\textbf{AQI\_ORI}}
            & MAE & 0.604 & 0.624 & 0.608 & 0.627 & \mysecond{0.598} & 0.621 & 0.858 & 0.798 & 0.795 & 0.648 & \myfirst{0.555} \\
            & MSE & 0.804 & 0.843 & 0.818 & 0.844 & \mysecond{0.741} & 0.843 & 1.448 & 1.438 & 1.313 & 0.925 & \myfirst{0.663} \\
            \midrule
            \multirow{2}{*}{\textbf{AQI\_IMP}}
            & MAE & \mysecond{0.650} & 0.665 & 0.668 & 0.666 & 0.653 & 0.663 & 0.941 & 0.946 & 0.857 & 0.733 & \myfirst{0.616} \\
            & MSE & 0.966 & 0.986 & 1.012 & 0.993 & \mysecond{0.893} & 0.995 & 1.775 & 1.928 & 1.543 & 1.206 & \myfirst{0.844} \\
            \bottomrule
        \end{tabular}%
    }
\end{table}

\textbf{Obs. \ding{184}: Imputation-based strategies consistently underperform direct prediction on naturally missing data.}
As shown in~\cref{tab:aqi_results_v2}, models trained on the fully imputed dataset~(AQI\_IMP) yield higher error rates on observed entries compared to the original sparse data~(AQI\_ORI). 
We attribute this degradation to distributional perturbations introduced by unsupervised imputation. 
Lacking ground truth, the imputer generates synthetic artifacts that deviate from real dynamics. 
Consequently, the downstream model is forced to fit these erratic perturbations, which distorts feature representation and impairs its ability to capture authentic dependencies in valid observations.
More implementation details are in App.~\ref{app:Natural_Missingness_Evaluations}.

\subsection{Ablation and Sensitivity Study}

\begin{figure*}[t]
    \centering
    \subfloat[Ablation Experiment Results]{ \includegraphics[width=0.28\textwidth]{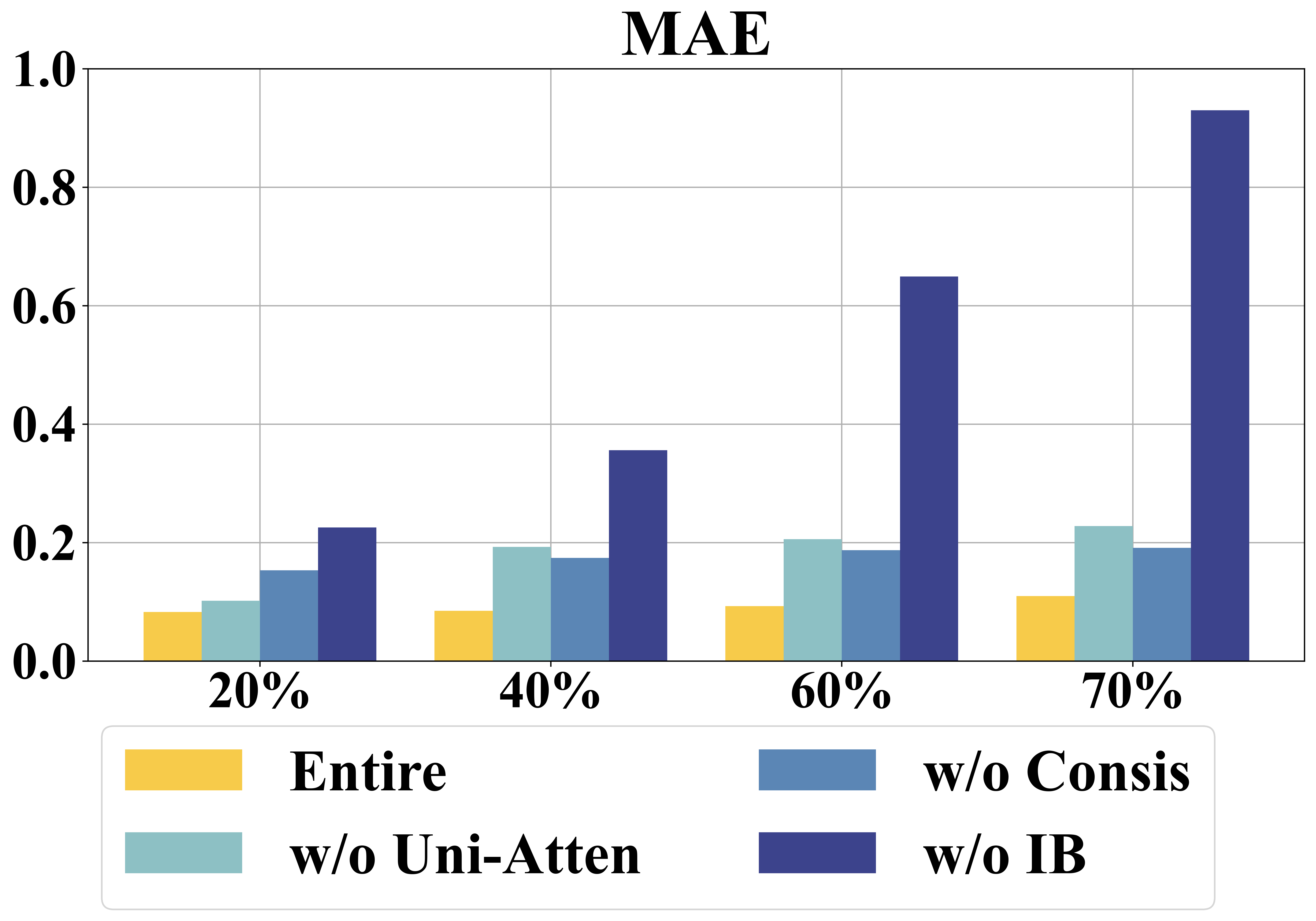} }
    \hfill
    \subfloat[Sensitivity Study Results]{ \includegraphics[width=0.68\textwidth]{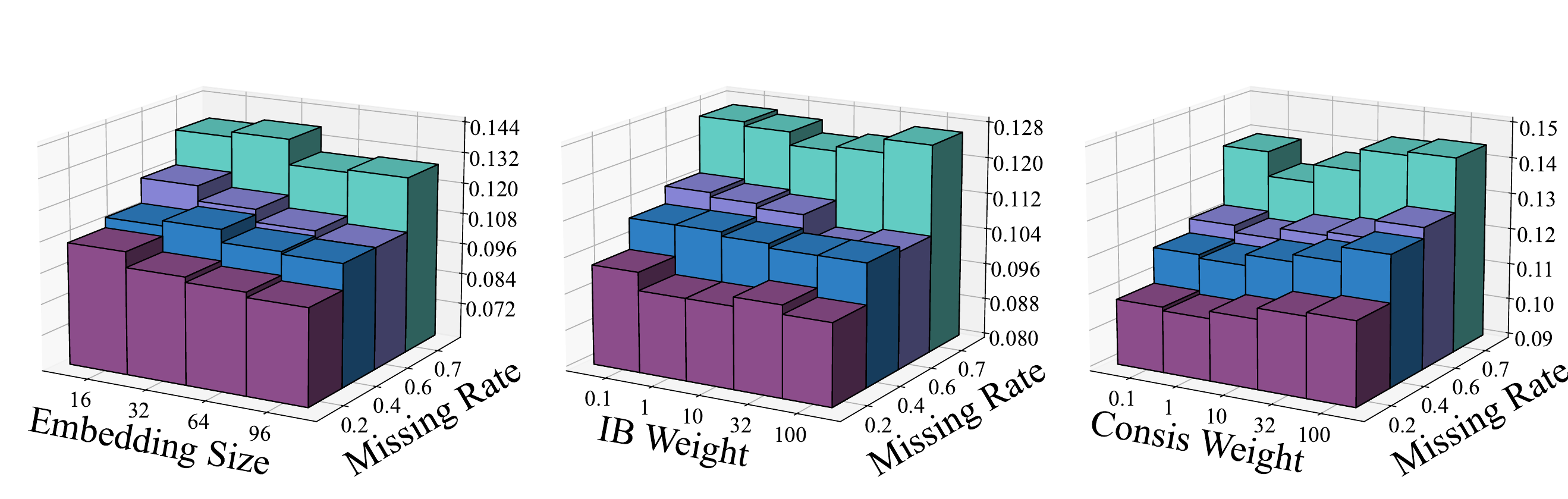} }
    \caption{Ablation and Sensitivity experiment results on PEMS-BAY dataset of
    \MODEL.}
    \label{fig:AblationAndSensitivity}
\end{figure*}

\renewcommand{\arraystretch}{1.25}
\begin{table*}
    [htbp]
    \begin{small}
        \centering
        \caption{Ablation study of consistency regularization under different
        missing rates on ETTh1.}
        \label{tab:consis_ablation_var} \resizebox{\textwidth}{!}{
        \begin{tabular}{l|cc|cc|cc|cc}
            \toprule \multirow{2}{*}{\textbf{Method}} & \multicolumn{2}{c|}{\textbf{Missing 20\%}} & \multicolumn{2}{c|}{\textbf{Missing 40\%}} & \multicolumn{2}{c|}{\textbf{Missing 60\%}} & \multicolumn{2}{c}{\textbf{Missing 70\%}} \\
                                                      & MAE                                        & MSE                                        & MAE                                        & MSE                                      & MAE                       & MSE                       & MAE                       & MSE                       \\
            \midrule w/o Consis                       & 0.235$\pm$0.0022                           & 0.264$\pm$0.0001                           & 0.283$\pm$0.0011                           & 0.276$\pm$0.0003                         & 0.339$\pm$0.0021          & 0.405$\pm$0.0003          & 0.448$\pm$0.0020          & 0.574$\pm$0.0010          \\
            \textbf{\MODEL}                          & \textbf{0.220$\pm$0.0001}                  & \textbf{0.171$\pm$0.0001}                  & \textbf{0.251$\pm$0.0001}                  & \textbf{0.249$\pm$0.0001}                & \textbf{0.267$\pm$0.0001} & \textbf{0.296$\pm$0.0001} & \textbf{0.288$\pm$0.0001} & \textbf{0.361$\pm$0.0008} \\
            \bottomrule
        \end{tabular}
        }
    \end{small}
\end{table*}


We further conduct ablation and parameter sensitivity studies on the PEMS-BAY, ETTh1, Electricity and Metr-LA
datasets under four missing rates to examine the contribution and robustness of
each component in \MODEL.
In the \textbf{Ablation Study}~(\cref{fig:AblationAndSensitivity}~(a) and App.~\ref{app:Extra_Ablation_Study}),
we design three ablation experiments with configurations as follows: (1) \textbf{w/o
Uni-Atten}: we replace the unified-variate attention mechanism with the vanilla
attention mechanism. (2) \textbf{w/o Consis}: we remove the consistency regularization.
(3) \textbf{w/o IB}: we remove the compactness and informativeness guidance of
IB. 
We get observations as follows:

\textbf{Obs. \ding{185}: Capturing variate correlations and ensuring consistency
are critical for direct forecasting.} 
Removing
either the unified-variate attention mechanism or the consistency regularization
leads to clear performance degradation across all missing settings.
This indicates that explicitly modeling inter-variate correlations and enforcing
representation consistency are both essential for accurate forecasting under
missingness.
Moreover, as shown in \cref{tab:consis_ablation_var}, 
consistency regularization plays an important role in stabilizing
training and reducing prediction variance, as reflected by the increased error
and instability observed when it is removed.

\textbf{Obs. \ding{186}: The Information Bottleneck principle is the model's foundational
component.} 
Among all ablation variants, eliminating the IB guidance
results in the most severe performance drop.
Compared with other components, the IB principle has a more fundamental impact on
model behavior, as it directly governs the balance between filtering irrelevant
noise and preserving predictive information from incomplete inputs.
The sharp degradation of the w/o IB variant, contrasted with the relatively stable
performance of other ablations, confirms that IB serves as the core mechanism that
enables \MODEL to learn robust and compact representations under missing values.

We additionally conduct the \textbf{Sensitivity Study}~(\cref{fig:AblationAndSensitivity}~(b) and App.~\ref{app:Extra_Sensitivity_Study}) of key hyperparameters, we
vary the weights assigned to the \textbf{Embedding Size}, \textbf{IB weight: $\alpha$},
and \textbf{Consis Weight: $\gamma$} to study how each impacts model performance.

\textbf{Obs. \ding{187}: \MODEL is robust to hyperparameter variations, though
over-regularization can be detrimental under high missing rates.} As shown in \cref{fig:AblationAndSensitivity}~(b),
a larger embedding size generally correlates with better performance. However,
the model remains effective even with a small embedding size~(e.g., 32), demonstrating
its efficiency in terms of computational and memory costs. For the IB and
consistency regularization weights, we observe a trade-off. At low missing rates,
higher weight values can improve accuracy. However, as the missing rate increases,
excessively high weights tend to over-regularize the model, which can hinder its
ability to capture complex variate correlations and thus degrade the final forecasting
performance.

\section{Related Work}

\paragraph{Multivariate Time Series Forecasting with Missing Values}
Existing MTSF methods~\citep{liu2023itransformer,wang2024timexer, timealign}, which typically assume complete data, suffer significant performance degradation when applied to partially observed datasets. To address this issue, research on MTSF-M has emerged, focusing mainly on two directions: two-stage frameworks and end-to-end models. 
Two-stage methods combine imputation models~\citep{cao2018brits, cini2021filling, marisca2022learning} with forecasting models~\citep{liu2023itransformer, wu2021autoformer, tashiro2021csdi}. However, this decoupled design often leads to error propagation across stages~\citep{chen2023biased}, reducing overall forecasting accuracy.
End-to-end approaches, on the other hand, aim to jointly impute missing values and perform forecasting by interleaving spatial and temporal modules~\citep{yu2024ginar}.
Despite their promise, these methods face a key limitation: the lack of ground truth for the missing values. As a result, the imputation process becomes noisy, which negatively impacts prediction performance.
To address these limitations, we propose a direct prediction method \MODEL, which integrates an IB-based Consistency Regularization to effectively identify relevant signals while filtering out redundant or noisy information, leading to more accurate forecasts.

\paragraph{Information Bottleneck for Time Series}
The IB principle 
provides a theoretical framework for learning compact representations that retain maximal task-relevant information while discarding irrelevant variability~\citep{tishby2000information}. 
In the context of time series modeling, IB is commonly implemented via Variational Autoencoders (VAEs)~\citep{kingma2013auto, voloshynovskiy2019information}.
Existing methods like GP-VAE~\citep{fortuin2020gp}, MTS-IB~\citep{ullmann2023multivariate}, and RIB~\citep{xu2018time} use the IB framework to model temporal dynamics.
However, these approaches face a key limitation: a direct application of the IB principle can cause the model to concentrate too narrowly on observed features~\citep{choi2023conditional, zhang2025loss}, thereby neglecting the broader variate correlations crucial for forecasting from incomplete data.
In contrast to these works, our proposed \MODEL applies the IB principle with a unified-attention mechanism and a consistency regularization.
This design enables the model to leverage the regularization benefits of IB while avoiding excessive information collapse, resulting in representations that are both compact and robust to sparsity in the inputs.

\section{Conclusion}

In this paper, we analyze the dominant `imputation-then-prediction' paradigm for MTSF-M tasks.
Our empirical analysis reveals a fundamental flaw in this framework: without direct supervision, imputation can corrupt data distribution and degrade, rather than improve, final forecasting accuracy. 
To address this, we propose a direct prediction paradigm and introduce \MODEL, a novel framework designed to learn directly from incomplete data. 
By leveraging the IB principle with unified-variate attention and consistency regularization, \MODEL effectively filters noise while capturing robust predictive signals from partial observations. Extensive experiments validate our method, showing that \MODEL achieves a significant 18\% improvement and confirms the superiority of direct prediction.

\section*{Impact Statement}

This paper presents work whose goal is to advance the field of Machine
Learning. There are many potential societal consequences of our work, none of
which we feel must be specifically highlighted here.


\bibliography{reference}
\bibliographystyle{icml2026}

\newpage
\appendix
\onecolumn


\section{Full Derivation}
\label{app:FullDerivation}

We illustrate the full derivation of the two terms of IB as follows.

\paragraph{Compactness Principle:}
\begin{equation}
    \begin{aligned}
        I_{\theta}(Z;X^{\text{o}})
        &=\mathbb{E}_{p(x^{\text{o}},z)}[\log \frac{p(x^{\text{o}},z)}{p(z)\cdot p(x^{\text{o}})}],\\
        &=\mathbb{E}_{p(x^{\text{o}},z)}[\log \frac{p(z|x^{\text{o}})\cdot p(x^{\text{o}})}{p(z)\cdot p(x^{\text{o}})}],\\
        &=\mathbb{E}_{p(x^{\text{o}},z)}[\log \frac{p(z|x^{\text{o}})}{p(z)}],\\
        &=\mathbb{E}_{p(x^{\text{o}},z)}[\log \frac{p(z|x^{\text{o}})}{p(z)}\cdot \frac{q(z)}{q(z)} ],\\
        &=\mathbb{E}_{p(x^{\text{o}},z)}[\log \frac{p(z|x^{\text{o}})}{q(z)}]-\mathbb{E}_{p(x^{\text{o}},z)}[\log \frac{p(z)}{q(z)}], \\
        &=\mathbb{E}_{p(x^{\text{o}},z)}[\log \frac{p(z|x^{\text{o}})}{q(z)}]-D_{KL}[p(z)\ ||\ q(z)], \\
        &=\mathbb{E}_{p(x^{\text{o}})}[D_{KL}(p(z|x^{\text{o}})||q(z))]-D_{KL}[p(z)\ ||\ q(z)], \\
        &\le \mathbb{E}_{p(x^{\text{o}})}[D_{KL}(p(z|x^{\text{o}})||p(z))].
    \end{aligned}
    \label{eq:full_loss_comp}
\end{equation}

\paragraph{Informativeness Principle:}
\begin{equation}
    \begin{aligned}
        I_{\theta}(Y;Z)=&\mathbb{E}_{p(z,y)}[\log \frac{p(z,y)}{p(z)\cdot p(y)}], \\
        =&\mathbb{E}_{p(z,y)}[\log \frac{p(y|z)\cdot p(z)}{p(y)\cdot p(z)}], \\
        =&\mathbb{E}_{p(z,y)}[\log \frac{p(y|z)}{p(y)}], \\
        =&\mathbb{E}_{p(z,y)}[\log \frac{p(y|z)\cdot q_{\theta}(y|z)}{p(y)\cdot q_{\theta}(y|z)}], \\
        =&\mathbb{E}_{p(z,y)}[\log \frac{q_{\theta}(y|z)}{p(y)}]+\mathbb{E}_{p(z,y)}[\log \frac{p(y|z)}{q_{\theta}(y|z)}], \\
        =&\mathbb{E}_{p(z,y)}[\log \frac{q_{\theta}(y|z)}{p(y)}] + \iint_{z,y} p(z)\cdot p(y|z)\cdot \log \frac{p(y|z)}{q_{\theta}(y|z)}\ dz dy,\\
        =&\mathbb{E}_{p(z,y)}[\log \frac{q_{\theta}(y|z)}{p(y)}] + \int_{z} p(z)\cdot D_{KL}[p(y|z)\ ||\ q_{\theta}(y|z)]\ dz\\
        \ge & \mathbb{E}_{p(z,y)}[\log \frac{q_{\theta}(y|z)}{p(y)}],\\
        =&\mathbb{E}_{p(z,y)}[\log q_{\theta}(y|z)]+H(Y),\\
        \ge & \mathbb{E}_{p(z,y)}[\log q_{\theta}(y|z)].\\
    \end{aligned}
    \label{eq:full_loss_pred}
\end{equation}
The inequalities of the upper and lower bound in~\cref{eq:full_loss_comp,eq:full_loss_pred} follow directly from the non-negativity of the KL-divergence and Entropy.

\section{Quantitative Analysis of Introduction}
\label{app:Quantitative_Analysis_of_Introduction}

\begin{table}[htbp]
    \centering
    \caption{Performance comparison on PEMS-BAY dataset~(40\% Missing).}
    \label{tab:intro_pems_bay_results}
    
    \begin{tabular}{lcccccc}
        \toprule
        \textbf{Metric} & \textbf{\MODEL} & \textbf{BiTGraph} & \textbf{DLinear} & \makecell[c]{\textbf{TimesNet} \\ \textbf{+ DLinear}} & \textbf{TimeXer} & \makecell[c]{\textbf{TimesNet} \\ \textbf{+ TimeXer}} \\
        \midrule
        MAE & \myfirst{0.093} & 0.413 & 0.156 & 0.148 & \mysecond{0.125} & 0.135 \\
        MSE & \myfirst{0.043} & 0.788 & 0.087 & 0.081 & \mysecond{0.051} & 0.073 \\
        \bottomrule
    \end{tabular}
\end{table}

\begin{itemize}[leftmargin=*]
    \item[\ding{182}] \textbf{Detrimental Imputation:} We demonstrate that imputation without ground truth is often harmful. 
    As visualized in~\cref{fig:intro_analysis}, methods following the ``imputation-then-prediction" paradigm fail to recover the true data distribution and instead reinforce biased patterns from partial observations.
    
    \item[\ding{183}] \textbf{Performance Degradation:} We provide counter-intuitive evidence that imputation actively harms prediction accuracy. 
    For instance, equipping the predictor TimeXer with the SOTA imputer TimesNet increases the MAE from 0.125 to 0.135 as shown in~\cref{tab:intro_pems_bay_results}.
    It also has no help in understanding the data distribution and variate correlations as demonstrated in~\cref{fig:intro_analysis}.

\end{itemize}

\section{Datasets}
\label{app:Datasets}

To assess the model's effectiveness and robustness in handling missing values,
we introduce synthetic missingness by randomly removing data points at varying
missing rates of 20\%, 40\%, 60\%, and 70\% with three different missing patterns.
During the experiments, we normalized the data to facilitate better model fitting.

We introduce information about datasets in~\cref{tab:dataset_stats_rebuttal}.

\begin{table*}[htbp]
    \centering
    \caption{Statistics of the 11 real-world datasets used in our experiments.}
    \label{tab:dataset_stats_rebuttal}
    
    \setlength{\tabcolsep}{4pt} 
    
    \resizebox{\textwidth}{!}{%
        \begin{tabular}{lccccccccccc}
            \toprule
            \textbf{Statistics} & \textbf{ETTh1} & \textbf{ETTh2} & \textbf{ETTm1} & \textbf{ETTm2} & \textbf{Electricity} & \textbf{PEMS-BAY} & \textbf{Metr-LA} & \textbf{BeijingAir} & \textbf{Weather} & \textbf{Exchange} & \textbf{AQI} \\
            \midrule
            \textbf{Time Steps} & 17,420 & 17,420 & 69,680 & 69,680 & 26,304 & 52,116 & 34,272 & 36,000 & 52,696 & 7,588 & 8759 \\
            \textbf{Variates}   & 7 & 7 & 7 & 7 & 321 & 325 & 207 & 7 & 21 & 8 & 36 \\
            \bottomrule
        \end{tabular}%
    }
\end{table*}

\section{Implementation Details}

\subsection{Comparison Evaluations}
\label{app:Comparison_Evaluations}
We use Adam optimizer \citep{kingma2014adam} to learn the parameters of all
models with $10^{-3}$ learning rate. The unified-variate attention of \MODEL is configured
with 2 layers and 4 heads, while the predictor is implemented as a simple 2-layer
MLP. Both historical and future time window sizes are set to 24 for all methods,
following the setting of BiTGraph~\citep{chen2023biased}. The patch length is
set to 8, so every time series in a time window is patched into three tokens.
The entire dataset is divided into training, validation, and testing sets with
ratios of 60\%, 20\%, and 20\%.
All methods are trained on an Nvidia 4090 GPU with PyTorch 2.0.
Hyperparameters of all baselines are consistent with their original papers.

\subsection{Natural Missingness Evaluations}
\label{app:Natural_Missingness_Evaluations}
To fairly compare direct prediction with the imputation-then-prediction paradigm, we first pre-trained TimesNet~\citep{wu2022timesnet} on the AQI training set for imputation by applying an additional 10\% point-missing mask on observed entries, matching the dataset’s estimated natural missing rate. 
We then used the trained TimesNet to impute all NaNs in AQI and trained/evaluated downstream forecasting models on this fully imputed dataset (AQI\_IMP). In contrast, we trained/evaluated the same models directly on the original AQI with natural missingness (AQI\_ORI).

\subsection{Training Cost}
\label{app:Training_Cost}
To ensure fair comparisons, we train all baseline models from scratch using identical dataset splits and experimental protocols, as detailed in~\cref{subsubsec:Implementation_Details}. 
We evaluate computational efficiency by reporting the memory footprint and parameter counts of every model on ETTh1 as follows.

\paragraph{Conclusion:} The results demonstrate that \MODEL maintains a computational cost comparable to efficient Transformer baselines (e.g., PatchTST~\citep{nie2022time}) while being more lightweight than complex methods such as CSDI~\citep{tashiro2021csdi} and TimesNet~\citep{wu2022timesnet}.

\begin{table}[htbp]
    \centering
    \caption{Comparison of model efficiency in terms of parameter count and memory cost. \MODEL achieves a balanced trade-off between performance and efficiency.}
    \label{tab:rebuttal_train_cost}
    \begin{tabular}{lrr}
        \toprule
        \textbf{Model} & \textbf{Parameters} & \textbf{Memory~(MB)} \\
        \midrule
        \textbf{\MODEL~(Ours)} & \textbf{37,450} & \textbf{148.03} \\
        \midrule
        DLinear & 1,200 & 18.99 \\
        PAttn & 15,640 & 55.13 \\
        SegRNN & 8,056 & 30.17 \\
        Transformer & 57,063 & 189.38 \\
        iTransformer & 39,768 & 154.18 \\
        PatchTST & 41,528 & 179.80 \\
        TSMixer & 6,837 & 21.16 \\
        WPMixer & 44,370 & 50.16 \\
        \midrule
        CSDI & 239,649 & 1,269.72 \\
        NeuralCDE & 37,767 & 41.40 \\
        ImputeFormer & 264,193 & 1,488.22 \\
        TimesNet & 863,895 & 1,427.44 \\
        \bottomrule
    \end{tabular}
\end{table}

\section{Extra Experiments}
\label{app:Extra_Exp}

\begin{figure*}[htbp]
    \centering
    \includegraphics[width=1\textwidth]{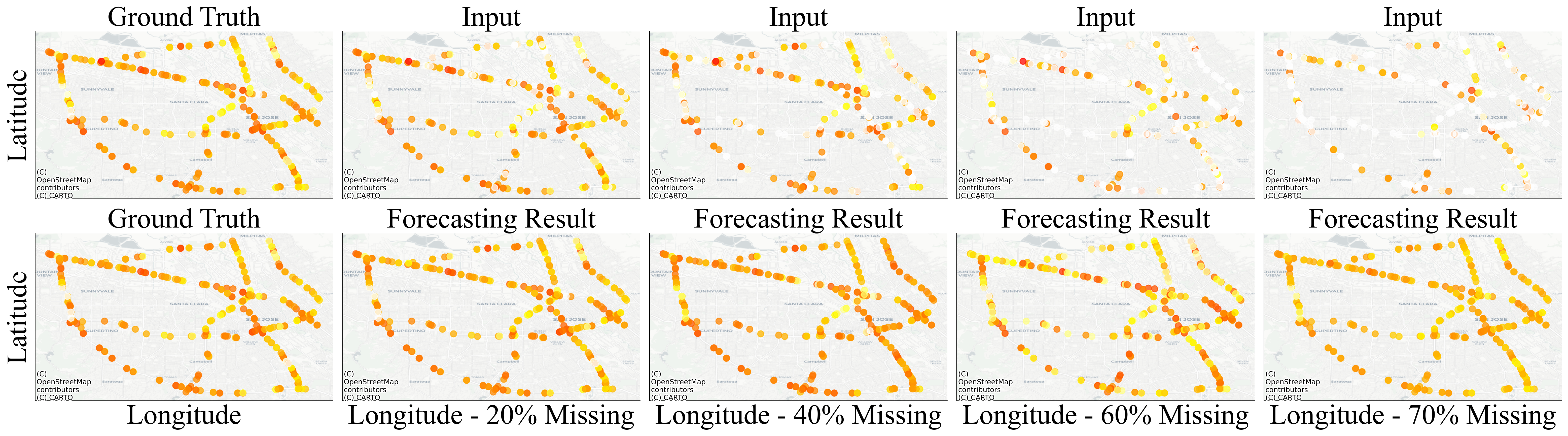}
    \caption{
    Visualization of the input and forecasting results of \MODEL on the PEMS-BAY dataset with missing rates from 20\% to 70\%.
    }
    \label{fig:Visualization}
\end{figure*}

\begin{figure*}[htbp]
    \centering
    \includegraphics[width=1\textwidth]{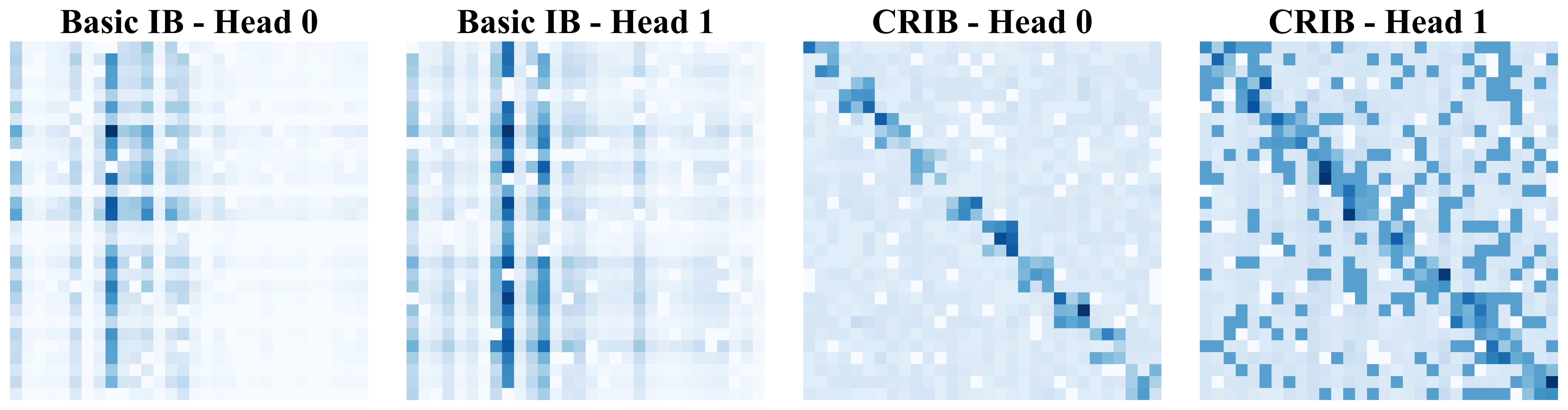}
    \caption{
    Visualization comparison of attention maps on the Metr-LA dataset with 60\% missing values. \textbf{Left}: Two attention maps of the direct application of IB on the standard Transformer.
    \textbf{Right}: Two attention maps of \MODEL.}
    \label{fig:Visualization-Attn}
\end{figure*}


\subsection{Forecasting Results Visualization}

We present a spatial visualization of forecasting results to demonstrate the effectiveness of \MODEL under varying missing rates. 
\cref{fig:Visualization} shows the final timestamp in the historical time window and the first forecasting timestamp on the PEMS-BAY dataset.
At lower missing rates~(20\% and 40\%), by effectively leveraging inter-variate correlations extracted from the data, \MODEL accurately predicts the future values.
Even at higher missing rates~(60\% and 70\%), \MODEL can maintain stable performance and predict the spatial distribution of the PEMS-BAY datasets.
These findings underscore \MODEL's capability to handle incomplete data and produce reliable predictions.

\subsection{Unified-Variate Attention Maps Visualization}
\label{app:IB-Attention}

In \cref{fig:Visualization-Attn}, we compare visualizations of directly applying IB on the Transformer with our proposed \MODEL. 
In the first experiment, a transformer model serves as the predictor. 
The \textbf{left} two figures clearly show that directly applying IB to the model would force the model to focus on a few specific values~(straight line attention), thereby neglecting global information. 
In contrast, the \textbf{right} figures reveal that \MODEL can not only capture the original intra-variate temporal correlations in one attention head but also effectively uncovers cross-variate correlations in another, rather than relying solely on raw correlations. 
As a result, the final forecasting performance is improved remarkably by our unified-variate attention mechanism and consistency regularization scheme.

\subsection{Experiments on Various Missing Patterns}
\label{app:Full_Exp_Missing_Pattern}

\begin{figure*}[htbp]
    \centering
    \includegraphics[width=1\textwidth]{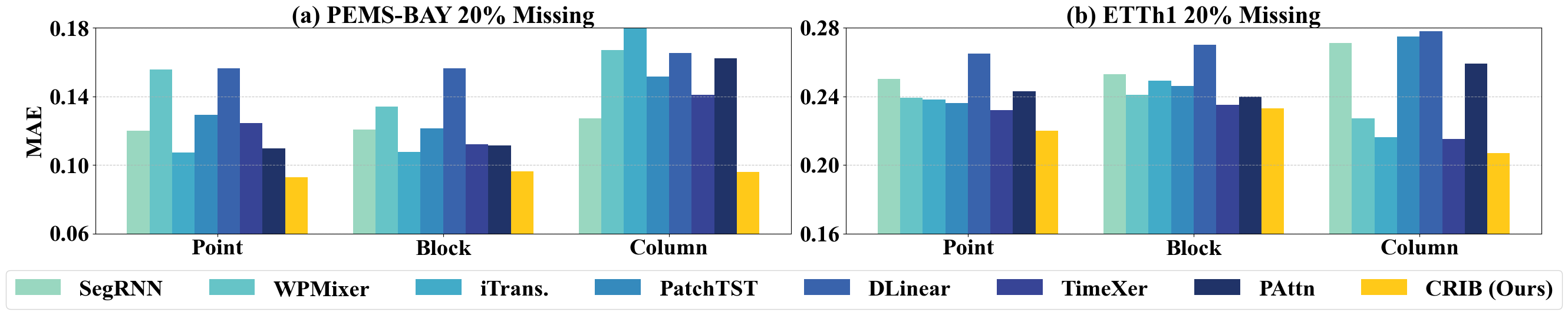}
    \caption{
    MAE comparison on PEMS-BAY and ETTh1 with point, block, and column missing patterns on 20\% missing rate.
    }
    \label{fig:mae_20}
\end{figure*}

\begin{figure*}[htbp]
    \centering
    \includegraphics[width=1\textwidth]{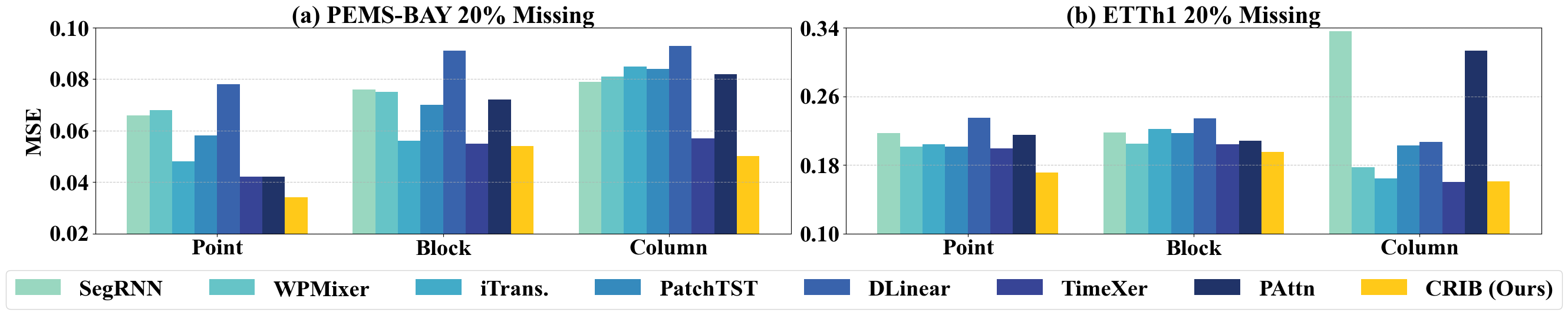}
    \caption{
    MSE comparison on PEMS-BAY and ETTh1 with point, block, and column missing patterns on 20\% missing rate.
    }
    \label{fig:mse_20}
\end{figure*}

\begin{figure*}[htbp]
    \centering
    \includegraphics[width=1\textwidth]{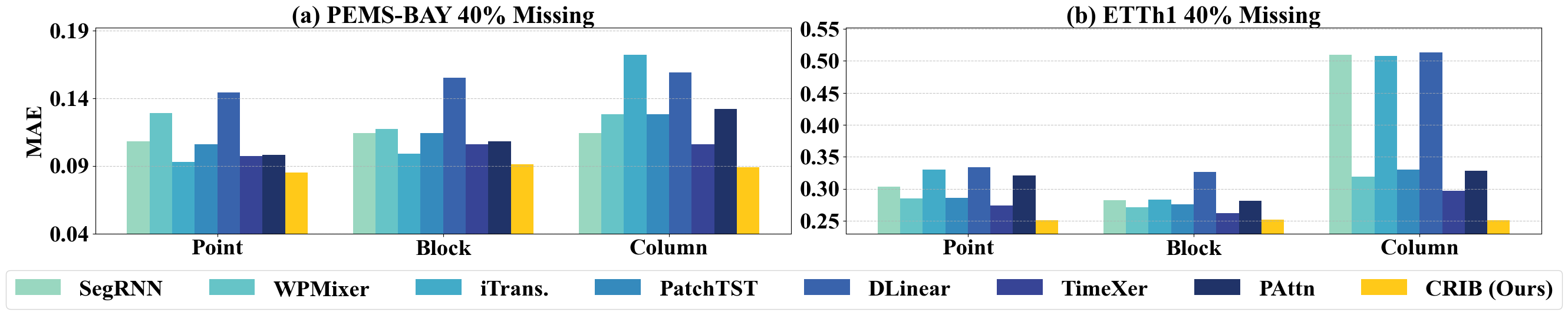}
    \caption{
    MAE comparison on PEMS-BAY and ETTh1 with point, block, and column missing patterns on 40\% missing rate.
    }
    \label{fig:mae_40}
\end{figure*}

\begin{figure*}[htbp]
    \centering
    \includegraphics[width=1\textwidth]{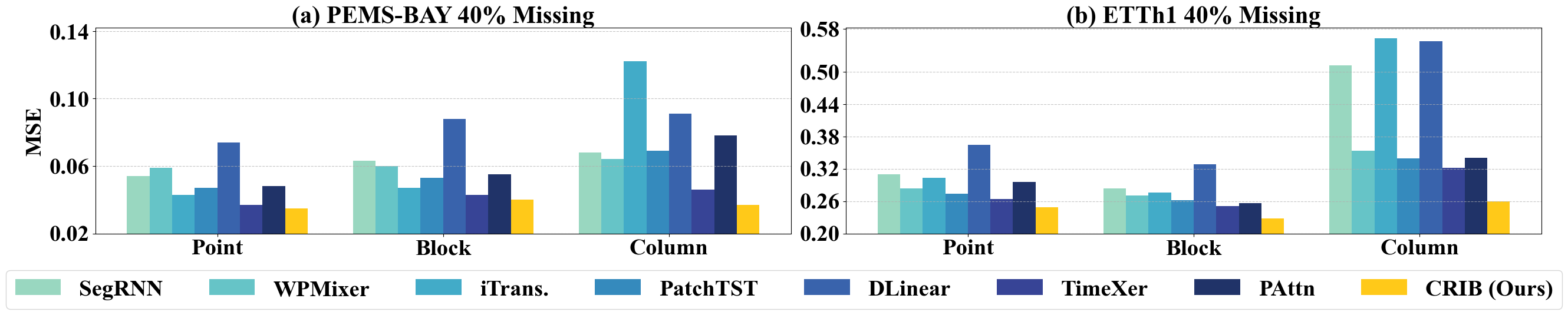}
    \caption{
    MSE comparison on PEMS-BAY and ETTh1 with point, block, and column missing patterns on 40\% missing rate.
    }
    \label{fig:mse_40}
\end{figure*}

\begin{figure*}[htbp]
    \centering
    \includegraphics[width=1\textwidth]{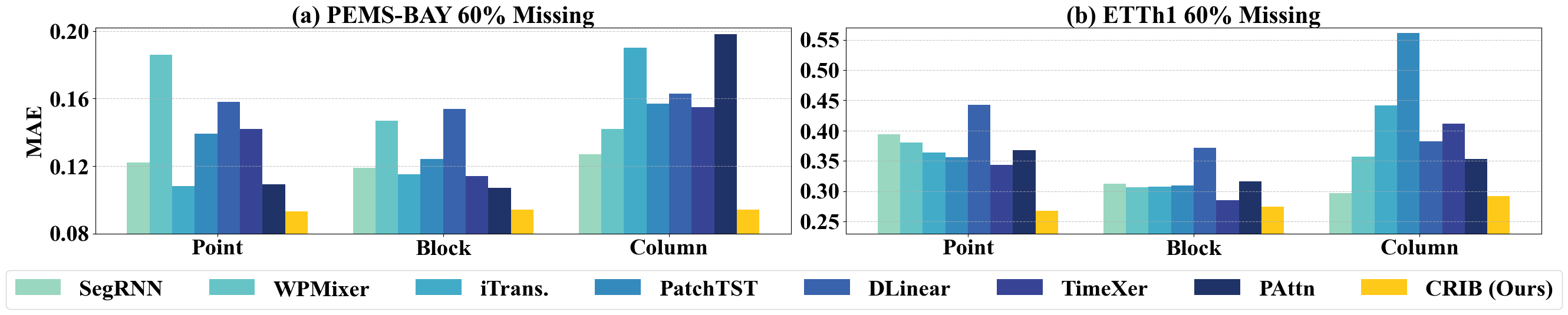}
    \caption{
    MAE comparison on PEMS-BAY and ETTh1 with point, block, and column missing patterns on 60\% missing rate.
    }
    \label{fig:mae_60}
\end{figure*}

\begin{figure*}[htbp]
    \centering
    \includegraphics[width=1\textwidth]{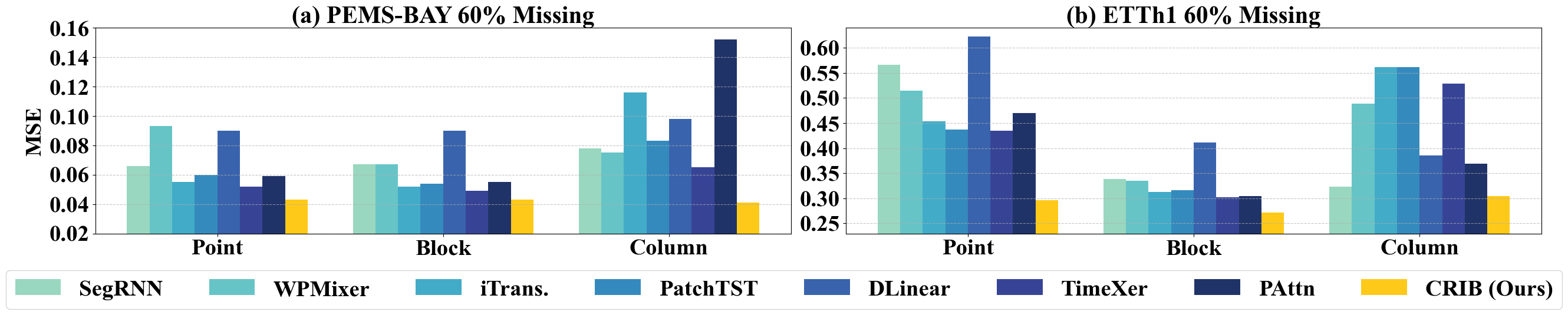}
    \caption{
    MSE comparison on PEMS-BAY and ETTh1 with point, block, and column missing patterns on 60\% missing rate.
    }
    \label{fig:mse_60}
\end{figure*}

\begin{figure*}[htbp]
    \centering
    \includegraphics[width=1\textwidth]{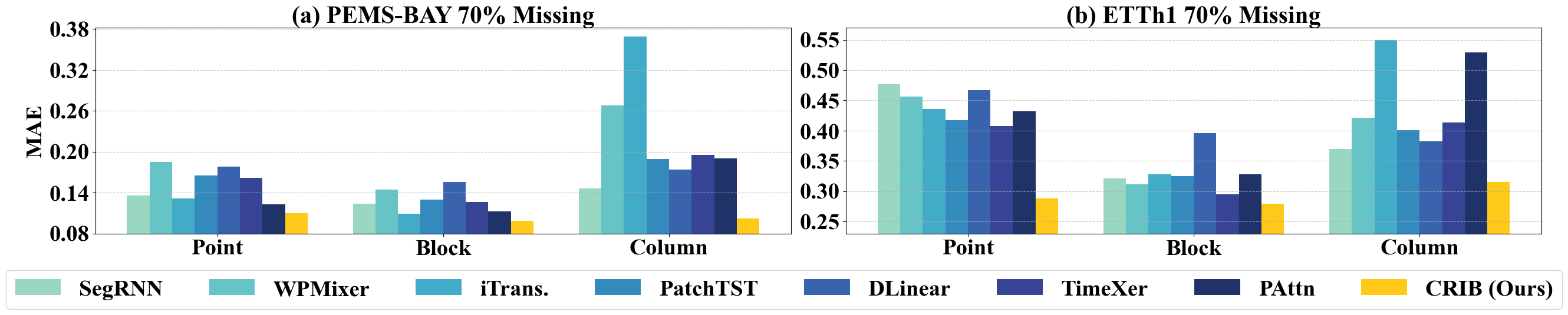}
    \caption{
    MAE comparison on PEMS-BAY and ETTh1 with point, block, and column missing patterns on 70\% missing rate.
    }
    \label{fig:mae_70}
\end{figure*}

\begin{figure*}[htbp]
    \centering
    \includegraphics[width=1\textwidth]{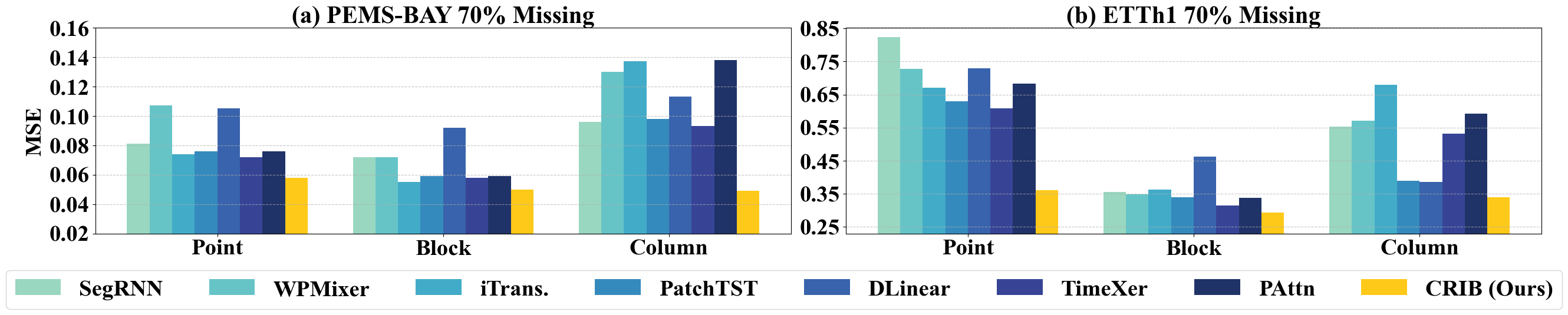}
    \caption{
    MSE comparison on PEMS-BAY and ETTh1 with point, block, and column missing patterns on 70\% missing rate.
    }
    \label{fig:mse_70}
\end{figure*}

\Cref{fig:mae_20,fig:mse_20,fig:mae_40,fig:mse_40,fig:mae_60,fig:mse_60,fig:mae_70,fig:mse_70} present the main forecasting results, comparing our proposed model, CRIB, against state-of-the-art baselines.
The results clearly show that CRIB consistently achieves the lowest MAE and MSE across all evaluated scenarios. 
This superiority holds true for both the PEMS-BAY and ETTh1 datasets, under point, block, and column missing patterns, and across a wide range of missing rates from 20\% to 70\%. 
Notably, while the performance of most baseline models degrades significantly as the missing rate increases, CRIB maintains its superior performance and stability.
This demonstrates the robustness and effectiveness of our direct-prediction approach, validating its superiority over existing methods, especially in challenging high-missing-rate environments.

\section{Extra Ablation Study}
\label{app:Extra_Ablation_Study}

We have done an extra ablation study on three cases of loss of \MODEL to prove its effectiveness. 
The ablation analysis across four datasets confirms the necessity of each component, as removing the Consistency Regularization~($\mathcal{L}_{\text{Consis}}$), Compactness~($\mathcal{L}_{\text{Comp}}$), or Informativeness~($\mathcal{L}_{\text{Pred}}$) objectives consistently leads to performance degradation, validating their collective role in ensuring robust and accurate forecasting in MTSF-M tasks.


\begin{table}[!ht]
    \centering
    \caption{Ablation study on ETTh1 dataset.}
    \label{tab:ablation_etth1}
    \setlength{\tabcolsep}{4pt} 
    \begin{tabular}{lcccccccc}
    \toprule
    \multirow{2}{*}{\textbf{ETTh1}} & \multicolumn{2}{c}{0.2} & \multicolumn{2}{c}{0.4} & \multicolumn{2}{c}{0.6} & \multicolumn{2}{c}{0.7} \\
    \cmidrule(lr){2-3} \cmidrule(lr){4-5} \cmidrule(lr){6-7} \cmidrule(lr){8-9}
     & MAE & MSE & MAE & MSE & MAE & MSE & MAE & MSE \\
    \midrule
    w/o Consis & 0.237 & 0.206 & 0.274 & 0.270 & 0.338 & 0.413 & 0.402 & 0.584 \\
    w/o Reg    & 0.236 & 0.206 & 0.274 & 0.270 & 0.339 & 0.410 & 0.400 & 0.579 \\
    w/o Pred   & 0.432 & 0.559 & 0.505 & 0.698 & 0.655 & 1.066 & 0.796 & 1.486 \\
    \midrule
    \textbf{Entire}     & \myfirst{0.220} & \myfirst{0.171} & \myfirst{0.251} & \myfirst{0.249} & \myfirst{0.267} & \myfirst{0.296} & \myfirst{0.288} & \myfirst{0.361} \\
    \bottomrule
    \end{tabular}
\end{table}


\begin{table}[!ht]
    \centering
    \caption{Ablation study on Elec dataset.}
    \label{tab:ablation_elec}
    \setlength{\tabcolsep}{4pt}
    \begin{tabular}{lcccccccc}
    \toprule
    \multirow{2}{*}{\textbf{Elec}} & \multicolumn{2}{c}{0.2} & \multicolumn{2}{c}{0.4} & \multicolumn{2}{c}{0.6} & \multicolumn{2}{c}{0.7} \\
    \cmidrule(lr){2-3} \cmidrule(lr){4-5} \cmidrule(lr){6-7} \cmidrule(lr){8-9}
     & MAE & MSE & MAE & MSE & MAE & MSE & MAE & MSE \\
    \midrule
    w/o Consis & 0.0201 & 0.0196 & 0.0267 & 0.0338 & 0.0346 & 0.0569 & 0.0415 & 0.0931 \\
    w/o Reg    & 0.0186 & 0.0175 & 0.0244 & 0.0286 & 0.0322 & 0.0544 & 0.0399 & 0.0980 \\
    w/o Pred   & 0.0881 & 0.5454 & 0.1243 & 0.9424 & 0.1818 & 1.8995 & 0.2283 & 2.9058 \\
    \midrule
    \textbf{Entire}     & \myfirst{0.0150} & \myfirst{0.0120} & \myfirst{0.0230} & \myfirst{0.0280} & \myfirst{0.0300} & \myfirst{0.0470} & \myfirst{0.0380} & \myfirst{0.0910} \\
    \bottomrule
    \end{tabular}
\end{table}


\begin{table}[!ht]
    \centering
    \caption{Ablation study on Metr-LA dataset.}
    \label{tab:ablation_metr}
    \setlength{\tabcolsep}{4pt}
    \begin{tabular}{lcccccccc}
    \toprule
    \multirow{2}{*}{\textbf{Metr-LA}} & \multicolumn{2}{c}{0.2} & \multicolumn{2}{c}{0.4} & \multicolumn{2}{c}{0.6} & \multicolumn{2}{c}{0.7} \\
    \cmidrule(lr){2-3} \cmidrule(lr){4-5} \cmidrule(lr){6-7} \cmidrule(lr){8-9}
     & MAE & MSE & MAE & MSE & MAE & MSE & MAE & MSE \\
    \midrule
    w/o Consis & 0.271 & 0.332 & 0.257 & 0.275 & 0.271 & 0.306 & 0.314 & 0.366 \\
    w/o Reg    & 0.253 & 0.307 & 0.251 & 0.275 & 0.266 & 0.307 & 0.311 & 0.364 \\
    w/o Pred   & 0.442 & 0.418 & 0.624 & 0.529 & 0.985 & 1.177 & 1.276 & 1.959 \\
    \midrule
    \textbf{Entire}     & \myfirst{0.248} & \myfirst{0.271} & \myfirst{0.249} & \myfirst{0.272} & \myfirst{0.265} & \myfirst{0.305} & \myfirst{0.309} & \myfirst{0.356} \\
    \bottomrule
    \end{tabular}
\end{table}


\begin{table}[!ht]
    \centering
    \caption{Ablation study on PEMS-BAY dataset.}
    \label{tab:ablation_pems}
    \setlength{\tabcolsep}{4pt}
    \begin{tabular}{lcccccccc}
    \toprule
    \multirow{2}{*}{\textbf{PEMS-BAY}} & \multicolumn{2}{c}{0.2} & \multicolumn{2}{c}{0.4} & \multicolumn{2}{c}{0.6} & \multicolumn{2}{c}{0.7} \\
    \cmidrule(lr){2-3} \cmidrule(lr){4-5} \cmidrule(lr){6-7} \cmidrule(lr){8-9}
     & MAE & MSE & MAE & MSE & MAE & MSE & MAE & MSE \\
    \midrule
    w/o Consis & 0.0959 & 0.0470 & 0.0882 & 0.0389 & 0.0976 & 0.0465 & 0.1296 & 0.0596 \\
    w/o Reg    & 0.0942 & 0.0453 & 0.0869 & 0.0373 & 0.0964 & 0.0456 & 0.1190 & 0.0591 \\
    w/o Pred   & 0.4006 & 0.2405 & 0.5618 & 0.3631 & 1.0832 & 1.2727 & 2.1480 & 4.8111 \\
    \midrule
    \textbf{Entire}     & \myfirst{0.0830} & \myfirst{0.0340} & \myfirst{0.0850} & \myfirst{0.0350} & \myfirst{0.0930} & \myfirst{0.0430} & \myfirst{0.1100} & \myfirst{0.0580} \\
    \bottomrule
    \end{tabular}
\end{table}

\section{Extra Sensitivity Study}
\label{app:Extra_Sensitivity_Study}

We analyze hyperparameter sensitivity in~\cref{fig:AblationAndSensitivity}~(b) and conduct additional sensitivity studies. 
Empirical results indicate that the optimal settings are consistent across different datasets and missing rates.
We set the weights for $\mathcal{L}_\text{Comp}$, $\mathcal{L}_\text{Pred}$, and $\mathcal{L}_\text{Consis}$ to $10^{-6}$, $1$, and $1$ as default, respectively.

\begin{table*}[htbp]
    \centering
    \caption{Sensitivity analysis of $\mathcal{L}_{\text{Pred}}$ weight across all missing rates.}
    \label{tab:sens_pred_horizontal}
    \setlength{\tabcolsep}{4pt}
    \resizebox{\textwidth}{!}{%
        \begin{tabular}{c|cc|cc|cc|cc|cc}
            \toprule
            \multirow{2}{*}{\textbf{Weight}} 
            & \multicolumn{2}{c|}{\textbf{0\% Missing}} 
            & \multicolumn{2}{c|}{\textbf{20\% Missing}} 
            & \multicolumn{2}{c|}{\textbf{40\% Missing}} 
            & \multicolumn{2}{c|}{\textbf{60\% Missing}} 
            & \multicolumn{2}{c}{\textbf{70\% Missing}} \\
            \cmidrule(lr){2-3} \cmidrule(lr){4-5} \cmidrule(lr){6-7} \cmidrule(lr){8-9} \cmidrule(lr){10-11}
             & \textbf{ETTh1} & \textbf{Exchange} 
             & \textbf{ETTh1} & \textbf{Exchange} 
             & \textbf{ETTh1} & \textbf{Exchange} 
             & \textbf{ETTh1} & \textbf{Exchange} 
             & \textbf{ETTh1} & \textbf{Exchange} \\
            \midrule
            0.1 & 0.201 & 0.0186 & 0.243 & 0.0332 & 0.281 & 0.0439 & 0.348 & 0.0413 & 0.412 & 0.0373 \\
            0.5 & 0.198 & 0.0186 & 0.238 & 0.0301 & 0.273 & 0.0294 & 0.340 & 0.0362 & 0.406 & 0.0423 \\
            1.0 & 0.198 & 0.0186 & 0.237 & 0.0287 & 0.274 & 0.0253 & 0.337 & 0.0363 & 0.403 & 0.0517 \\
            2.0 & 0.199 & 0.0186 & 0.236 & 0.0261 & 0.273 & 0.0261 & 0.339 & 0.0363 & 0.400 & 0.0475 \\
            5.0 & 0.200 & 0.0186 & 0.237 & 0.0246 & 0.273 & 0.0258 & 0.338 & 0.0287 & 0.401 & 0.0320 \\
            \bottomrule
        \end{tabular}%
    }
\end{table*}

\begin{table*}[htbp]
    \centering
    \caption{Sensitivity analysis of $\mathcal{L}_{\text{Comp}}$ weight across all missing rates.}
    \label{tab:sens_comp_horizontal}
    \setlength{\tabcolsep}{4pt}
    \resizebox{\textwidth}{!}{%
        \begin{tabular}{c|cc|cc|cc|cc|cc}
            \toprule
            \multirow{2}{*}{\textbf{Weight}} 
            & \multicolumn{2}{c|}{\textbf{0\% Missing}} 
            & \multicolumn{2}{c|}{\textbf{20\% Missing}} 
            & \multicolumn{2}{c|}{\textbf{40\% Missing}} 
            & \multicolumn{2}{c|}{\textbf{60\% Missing}} 
            & \multicolumn{2}{c}{\textbf{70\% Missing}} \\
            \cmidrule(lr){2-3} \cmidrule(lr){4-5} \cmidrule(lr){6-7} \cmidrule(lr){8-9} \cmidrule(lr){10-11}
             & \textbf{ETTh1} & \textbf{Exchange}
             & \textbf{ETTh1} & \textbf{Exchange}
             & \textbf{ETTh1} & \textbf{Exchange}
             & \textbf{ETTh1} & \textbf{Exchange}
             & \textbf{ETTh1} & \textbf{Exchange} \\
            \midrule
            10        & 0.214 & 0.0192 & 0.260 & 0.0509 & 0.314 & 0.0806 & 0.417 & 0.2131 & 0.506 & 0.2851 \\
            1         & 0.208 & 0.0188 & 0.251 & 0.0471 & 0.296 & 0.0835 & 0.365 & 0.1393 & 0.437 & 0.2293 \\
            $10^{-2}$ & 0.199 & 0.0186 & 0.238 & 0.0300 & 0.274 & 0.0319 & 0.341 & 0.0435 & 0.406 & 0.0499 \\
            $10^{-6}$ & 0.198 & 0.0186 & 0.237 & 0.0287 & 0.274 & 0.0253 & 0.337 & 0.0363 & 0.403 & 0.0517 \\
            $10^{-10}$& 0.199 & 0.0186 & 0.237 & 0.0276 & 0.273 & 0.0260 & 0.339 & 0.0417 & 0.403 & 0.0408 \\
            \bottomrule
        \end{tabular}%
    }
\end{table*}

\begin{table*}[htbp]
    \centering
    \caption{Sensitivity analysis of $\mathcal{L}_{\text{Consis}}$ weight across all missing rates.}
    \label{tab:sens_consis_horizontal}
    \setlength{\tabcolsep}{4pt}
    \resizebox{\textwidth}{!}{%
        \begin{tabular}{c|cc|cc|cc|cc|cc}
            \toprule
            \multirow{2}{*}{\textbf{Weight}} 
            & \multicolumn{2}{c|}{\textbf{0\% Missing}} 
            & \multicolumn{2}{c|}{\textbf{20\% Missing}} 
            & \multicolumn{2}{c|}{\textbf{40\% Missing}} 
            & \multicolumn{2}{c|}{\textbf{60\% Missing}} 
            & \multicolumn{2}{c}{\textbf{70\% Missing}} \\
            \cmidrule(lr){2-3} \cmidrule(lr){4-5} \cmidrule(lr){6-7} \cmidrule(lr){8-9} \cmidrule(lr){10-11}
             & \textbf{ETTh1} & \textbf{Exchange}
             & \textbf{ETTh1} & \textbf{Exchange}
             & \textbf{ETTh1} & \textbf{Exchange}
             & \textbf{ETTh1} & \textbf{Exchange}
             & \textbf{ETTh1} & \textbf{Exchange} \\
            \midrule
            0.1 & 0.201 & 0.0186 & 0.237 & 0.0231 & 0.274 & 0.0270 & 0.340 & 0.0301 & 0.400 & 0.0341 \\
            0.5 & 0.199 & 0.0186 & 0.236 & 0.0260 & 0.274 & 0.0273 & 0.339 & 0.0378 & 0.401 & 0.0368 \\
            1.0 & 0.198 & 0.0186 & 0.237 & 0.0287 & 0.274 & 0.0253 & 0.337 & 0.0363 & 0.403 & 0.0517 \\
            2.0 & 0.199 & 0.0186 & 0.239 & 0.0306 & 0.275 & 0.0300 & 0.340 & 0.0352 & 0.404 & 0.0427 \\
            5.0 & 0.200 & 0.0186 & 0.240 & 0.0312 & 0.278 & 0.0377 & 0.344 & 0.0376 & 0.410 & 0.0357 \\
            \bottomrule
        \end{tabular}%
    }
\end{table*}

\section{Full Experiments}
\label{app:Full_Exp}

\begin{table*}[htbp]
\centering
\caption{Performance comparison of different models for multivariate time series forecasting with missing values. 
Missing rate is set at 20\%, 40\%, 60\%, and 70\%.
The best results are highlighted in \myfirst{Bold} and the second-best is highlighted in \mysecond{Underline}.
}
\label{tab:MainExp_Transposed}
\begin{small}
\renewcommand{\multirowsetup}{\centering}
\setlength{\tabcolsep}{1pt}
\resizebox{\textwidth}{!}{
\begin{tabular}{cc|c|c|c|c|c|cc|cc|cc|cc|cc|cc|cc|c}
\toprule
\multirow{2}{*}{\textbf{\scalebox{0.83}{Data}}} 
& \multirow{2}{*}{\textbf{\scalebox{0.80}{Metric}}} 
& \multicolumn{1}{c}{\scalebox{0.81}{BiTGraph}} 
& \multicolumn{1}{c}{\scalebox{0.81}{BRITS}} 
& \multicolumn{1}{c}{\scalebox{0.81}{GRIN}} 
& \multicolumn{1}{c}{\scalebox{0.81}{SAITS}} 
& \multicolumn{1}{c}{\scalebox{0.81}{SPIN}} 
& \multicolumn{2}{c}{\scalebox{0.81}{SegRNN}} 
& \multicolumn{2}{c}{\scalebox{0.81}{WPMixer}} 
& \multicolumn{2}{c}{\scalebox{0.81}{iTransformer}} 
& \multicolumn{2}{c}{\scalebox{0.81}{PatchTST}} 
& \multicolumn{2}{c}{\scalebox{0.81}{DLinear}} 
& \multicolumn{2}{c}{\scalebox{0.81}{TimeXer}} 
& \multicolumn{2}{c}{\scalebox{0.81}{PAttn}} 
& \scalebox{0.81}{\textbf{Ours}} \\

\cmidrule(lr){3-3} \cmidrule(lr){4-4} \cmidrule(lr){5-5} \cmidrule(lr){6-6} \cmidrule(lr){7-7} \cmidrule(lr){8-9} \cmidrule(lr){10-11} \cmidrule(lr){12-13} \cmidrule(lr){14-15} \cmidrule(lr){16-17} \cmidrule(lr){18-19} \cmidrule(lr){20-21} \cmidrule(lr){22-22}

& & \multicolumn{1}{c}{\scalebox{0.81}{Original}} 
& \multicolumn{1}{c}{\scalebox{0.81}{Original}} 
& \multicolumn{1}{c}{\scalebox{0.81}{Imputed}}  
& \multicolumn{1}{c}{\scalebox{0.81}{Original}}  
& \multicolumn{1}{c}{\scalebox{0.81}{Imputed}}   
& \scalebox{0.81}{Original} & \scalebox{0.81}{Imputed} 
& \scalebox{0.81}{Original} & \scalebox{0.81}{Imputed} 
& \scalebox{0.81}{Original} & \scalebox{0.81}{Imputed} 
& \scalebox{0.81}{Original} & \scalebox{0.81}{Imputed} 
& \scalebox{0.81}{Original} & \scalebox{0.81}{Imputed} 
& \scalebox{0.81}{Original} & \scalebox{0.81}{Imputed} 
& \scalebox{0.81}{Original} & \scalebox{0.81}{Imputed} 
& \multicolumn{1}{c}{\scalebox{0.81}{Original}} \\   


\midrule
\multirow{8}{*}{\rotatebox{90}{\scalebox{0.99}{\textbf{PEMS-BAY}}}}
& \scalebox{0.80}{MAE@20\%} & 0.403 & 0.351 & 0.343 & OOM & 0.218 & 0.114 & 0.231 & 0.122 & 0.249 & \mysecond{0.097} & 0.153 & 0.107 & 0.158 & 0.145 & 0.163 & \mysecond{0.097} & 0.146 & 0.109 & 0.173 & \myfirst{0.083} \\
& \scalebox{0.80}{MSE@20\%} & 0.754 & 0.664 & 0.585 & OOM & 0.234 & 0.066 & 0.232 & 0.068 & 0.193 & 0.048 & 0.094 & 0.058 & 0.107 & 0.078 & 0.096 & \mysecond{0.042} & 0.087 & 0.062 & 0.111 & \myfirst{0.034} \\
\cmidrule(lr){2-2} \cmidrule(lr){3-22}
& \scalebox{0.80}{MAE@40\%} & 0.411 & 0.360 & 0.346 & OOM & 0.288 & 0.108 & 0.179 & 0.129 & 0.203 & \mysecond{0.093} & 0.127 & 0.106 & 0.142 & 0.144 & 0.138 & 0.097 & 0.140 & 0.098 & 0.165 & \myfirst{0.085} \\
& \scalebox{0.80}{MSE@40\%} & 0.777 & 0.696 & 0.609 & OOM & 0.360 & 0.054 & 0.185 & 0.059 & 0.135 & 0.043 & 0.074 & 0.047 & 0.087 & 0.074 & 0.069 & \mysecond{0.037} & 0.073 & 0.048 & 0.100 & \myfirst{0.035} \\
\cmidrule(lr){2-2} \cmidrule(lr){3-22}
& \scalebox{0.80}{MAE@60\%} & 0.419 & 0.372 & 0.355 & OOM & 0.501 & 0.122 & 0.153 & 0.186 & 0.181 & \mysecond{0.108} & 0.107 & 0.139 & 0.122 & 0.158 & 0.141 & 0.142 & 0.121 & 0.109 & 0.121 & \myfirst{0.093} \\
& \scalebox{0.80}{MSE@60\%} & 0.806 & 0.720 & 0.647 & OOM & 0.948 & 0.066 & 0.187 & 0.093 & 0.118 & 0.055 & 0.060 & 0.060 & 0.072 & 0.090 & 0.073 & \mysecond{0.052} & 0.061 & 0.059 & 0.073 & \myfirst{0.043} \\
\cmidrule(lr){2-2} \cmidrule(lr){3-22}
& \scalebox{0.80}{MAE@70\%} & 0.420 & 0.382 & 0.356 & OOM & 0.601 & 0.136 & 0.148 & 0.185 & 0.173 & 0.131 & 0.114 & 0.165 & 0.134 & 0.178 & 0.152 & 0.162 & 0.134 & \mysecond{0.123} & 0.131 & \myfirst{0.110} \\
& \scalebox{0.80}{MSE@70\%} & 0.816 & 0.742 & 0.653 & OOM & 1.053 & 0.081 & 0.208 & 0.107 & 0.116 & 0.074 & 0.061 & 0.076 & 0.079 & 0.105 & 0.084 & \mysecond{0.072} & 0.071 & 0.076 & 0.081 & \myfirst{0.058} \\

\midrule

\multirow{8}{*}{\rotatebox{90}{\scalebox{0.99}{\textbf{Metr-LA}}}}
& \scalebox{0.80}{MAE@20\%} & 0.435 & 0.351 & 0.387 & 0.484 & 0.336 & 0.280 & 0.356 & 0.300 & 0.372 & \mysecond{0.256} & 0.308 & 0.265 & 0.323 & 0.319 & 0.372 & 0.296 & 0.306 & 0.268 & 0.324 & \myfirst{0.248} \\
& \scalebox{0.80}{MSE@20\%} & 0.760 & 0.596 & 0.638 & 0.743 & 0.576 & 0.319 & 0.406 & 0.320 & 0.434 & 0.309 & 0.347 & 0.313 & 0.374 & 0.327 & 0.371 & \mysecond{0.303} & 0.345 & 0.323 & 0.370 & \myfirst{0.271} \\
\cmidrule(lr){2-2} \cmidrule(lr){3-22}
& \scalebox{0.80}{MAE@40\%} & 0.442 & 0.359 & 0.390 & 0.463 & 0.452 & 0.317 & 0.306 & 0.335 & 0.334 & \mysecond{0.254} & 0.280 & 0.302 & 0.307 & 0.346 & 0.344 & 0.293 & 0.291 & 0.308 & 0.282 & \myfirst{0.249} \\
& \scalebox{0.80}{MSE@40\%} & 0.756 & 0.600 & 0.667 & 0.697 & 0.692 & 0.305 & 0.321 & 0.301 & 0.349 & \mysecond{0.273} & 0.297 & 0.286 & 0.321 & 0.299 & 0.315 & \mysecond{0.273} & 0.306 & 0.305 & 0.308 & \myfirst{0.272} \\
\cmidrule(lr){2-2} \cmidrule(lr){3-22}
& \scalebox{0.80}{MAE@60\%} & 0.449 & 0.371 & 0.386 & 0.434 & 0.856 & 0.324 & 0.293 & 0.377 & 0.327 & \mysecond{0.274} & 0.280 & 0.341 & 0.296 & 0.426 & 0.360 & 0.327 & 0.295 & 0.309 & 0.277 & \myfirst{0.265} \\
& \scalebox{0.80}{MSE@60\%} & 0.760 & 0.615 & 0.649 & 0.719 & 1.196 & 0.326 & 0.334 & 0.357 & 0.357 & 0.309 & 0.314 & \mysecond{0.308} & 0.332 & 0.381 & 0.351 & 0.309 & 0.323 & 0.316 & 0.329 & \myfirst{0.305} \\
\cmidrule(lr){2-2} \cmidrule(lr){3-22}
& \scalebox{0.80}{MAE@70\%} & 0.452 & 0.382 & 0.393 & 0.422 & 0.856 & 0.351 & 0.301 & 0.411 & 0.334 & \mysecond{0.307} & 0.291 & 0.345 & 0.297 & 0.506 & 0.388 & 0.371 & 0.299 & 0.323 & 0.293 & \myfirst{0.309} \\
& \scalebox{0.80}{MSE@70\%} & 0.765 & 0.632 & 0.658 & 0.723 & 1.397 & 0.429 & 0.380 & 0.446 & 0.401 & 0.378 & 0.361 & 0.372 & 0.370 & 0.485 & 0.411 & 0.369 & 0.359 & 0.405 & 0.371 & \myfirst{0.356} \\

\midrule

\multirow{8}{*}{\rotatebox{90}{\scalebox{0.99}{\textbf{ETTh1}}}}
& \scalebox{0.80}{MAE@20\%} & 0.257 & \mysecond{0.232} & 0.234 & 0.369 & 0.232 & 0.250 & 0.394 & 0.239 & 0.386 & 0.238 & 0.417 & 0.236 & 0.398 & 0.265 & 0.538 & \mysecond{0.232} & 0.341 & 0.243 & 0.432 & \myfirst{0.220} \\
& \scalebox{0.80}{MSE@20\%} & 0.307 & 0.378 & 0.282 & 0.457 & \mysecond{0.191} & 0.217 & 0.331 & 0.201 & 0.323 & 0.204 & 0.374 & 0.201 & 0.382 & 0.235 & 0.508 & 0.199 & 0.296 & 0.215 & 0.387 & \myfirst{0.171} \\
\cmidrule(lr){2-2} \cmidrule(lr){3-22}
& \scalebox{0.80}{MAE@40\%} & 0.278 & 0.317 & 0.338 & 0.349 & 0.320 & 0.303 & 0.403 & 0.285 & 0.383 & 0.330 & 0.493 & 0.286 & 0.419 & 0.334 & 0.616 & \mysecond{0.274} & 0.340 & 0.321 & 0.506 & \myfirst{0.251} \\
& \scalebox{0.80}{MSE@40\%} & 0.316 & 0.373 & 0.386 & 0.430 & 0.346 & 0.310 & 0.384 & 0.284 & 0.352 & 0.303 & 0.556 & 0.274 & 0.445 & 0.365 & 0.646 & \mysecond{0.264} & 0.323 & 0.296 & 0.518 & \myfirst{0.249} \\
\cmidrule(lr){2-2} \cmidrule(lr){3-22}
& \scalebox{0.80}{MAE@60\%} & 0.394 & 0.432 & 0.399 & 0.380 & 0.598 & 0.394 & 0.445 & 0.380 & 0.404 & 0.364 & 0.382 & 0.356 & 0.355 & 0.443 & 0.631 & \mysecond{0.343} & 0.346 & 0.368 & 0.399 & \myfirst{0.267} \\
& \scalebox{0.80}{MSE@60\%} & 0.493 & 0.499 & 0.498 & 0.482 & 0.667 & 0.566 & 0.539 & 0.514 & 0.451 & 0.454 & 0.454 & 0.437 & 0.418 & 0.623 & 0.763 & \mysecond{0.435} & 0.393 & 0.470 & 0.462 & \myfirst{0.296} \\
\cmidrule(lr){2-2} \cmidrule(lr){3-22}
& \scalebox{0.80}{MAE@70\%} & 0.420 & 0.449 & 0.455 & \mysecond{0.390} & 0.599 & 0.477 & 0.459 & 0.456 & 0.424 & 0.436 & 0.384 & 0.417 & 0.373 & 0.567 & 0.608 & 0.408 & 0.363 & 0.432 & 0.390 & \myfirst{0.288} \\
& \scalebox{0.80}{MSE@70\%} & 0.432 & 0.436 & 0.433 & \mysecond{0.461} & 0.669 & 0.824 & 0.653 & 0.728 & 0.541 & 0.671 & 0.507 & 0.630 & 0.497 & 1.016 & 0.812 & 0.608 & 0.468 & 0.683 & 0.514 & \myfirst{0.361} \\

\midrule

\multirow{8}{*}{\rotatebox{90}{\scalebox{0.99}{\textbf{Electricity}}}}
& \scalebox{0.80}{MAE@20\%} & 0.029 & \mysecond{0.018} & 0.020 & 0.050 & 0.021 & 0.051 & 0.357 & 0.026 & 0.348 & 0.020 & 0.172 & 0.020 & 0.143 & 0.039 & 0.169 & 0.019 & 0.129 & 0.022 & 0.197 & \myfirst{0.015} \\
& \scalebox{0.80}{MSE@20\%} & 0.123 & 0.026 & 0.015 & 0.243 & 0.028 & 0.478 & 0.976 & 0.035 & 0.506 & 0.027 & 0.266 & 0.028 & 0.236 & 0.075 & 0.412 & \mysecond{0.022} & 0.158 & 0.027 & 0.499 & \myfirst{0.012} \\
\cmidrule(lr){2-2} \cmidrule(lr){3-22}
& \scalebox{0.80}{MAE@40\%} & 0.028 & 0.030 & 0.030 & 0.051 & 0.031 & 0.066 & 0.281 & 0.038 & 0.232 & 0.031 & 0.153 & 0.029 & 0.120 & 0.058 & 0.194 & \mysecond{0.024} & 0.089 & 0.042 & 0.188 & \myfirst{0.023} \\
& \scalebox{0.80}{MSE@40\%} & 0.116 & 0.054 & 0.066 & 0.266 & 0.070 & 0.722 & 1.072 & 0.083 & 0.267 & \mysecond{0.035} & 0.937 & 0.045 & 0.526 & 0.185 & 1.835 & 0.038 & 0.093 & 0.064 & 1.200 & \myfirst{0.028} \\
\cmidrule(lr){2-2} \cmidrule(lr){3-22}
& \scalebox{0.80}{MAE@60\%} & 0.038 & 0.044 & 0.041 & 0.053 & 0.223 & 0.089 & 0.210 & 0.056 & 0.159 & 0.041 & 0.118 & 0.040 & 0.089 & 0.086 & 0.228 & \mysecond{0.032} & 0.060 & 0.048 & 0.133 & \myfirst{0.030} \\
& \scalebox{0.80}{MSE@60\%} & 0.109 & 0.054 & \mysecond{0.059} & 0.258 & 0.667 & 1.185 & 1.412 & 0.197 & 0.184 & 0.065 & 0.700 & 0.110 & 0.496 & 0.465 & 2.348 & 0.062 & 0.073 & 0.139 & 1.248 & \myfirst{0.047} \\
\cmidrule(lr){2-2} \cmidrule(lr){3-22}
& \scalebox{0.80}{MAE@70\%} & 0.049 & 0.048 & 0.045 & 0.058 & 0.271 & 0.107 & 0.174 & 0.075 & 0.135 & 0.046 & 0.079 & 0.052 & 0.067 & 0.111 & 0.249 & \mysecond{0.041} & 0.055 & 0.056 & 0.090 & \myfirst{0.038} \\
& \scalebox{0.80}{MSE@70\%} & 0.104 & \mysecond{0.102} & 0.105 & 0.296 & 0.669 & 1.655 & 1.684 & 0.374 & 0.187 & 0.091 & 0.287 & 0.184 & 0.258 & 0.888 & 2.405 & 0.135 & 0.075 & 0.230 & 0.510 & \myfirst{0.091} \\
\bottomrule
\end{tabular}
}
\end{small}
\end{table*}

\begin{table*}[htbp]
\centering
\caption{Performance comparison on six additional datasets with varying missing rates.
The best results are highlighted in \myfirst{Bold}.
}
\label{tab:rebuttal_comp_results}
\renewcommand{\arraystretch}{0.85}
\begin{scriptsize}
\renewcommand{\multirowsetup}{\centering}
\setlength{\tabcolsep}{1pt}
\newcolumntype{C}{>{\centering\arraybackslash}p{1.05cm}}
\resizebox{\textwidth}{!}{
\begin{tabular}{cC|C|C|C|C|C|C|C|C|C|C|C}
\toprule
\textbf{\scalebox{0.75}{Data}}
& \textbf{\scalebox{0.75}{Metric}}
& \scalebox{0.75}{SegRNN}
& \scalebox{0.75}{WPMixer}
& \scalebox{0.65}{iTransformer}
& \scalebox{0.75}{PatchTST}
& \scalebox{0.75}{DLinear}
& \scalebox{0.75}{PAttn}
& \scalebox{0.75}{CSDI}
& \scalebox{0.65}{NeuralCDE}
& \scalebox{0.6}{ImputeFormer}
& \scalebox{0.75}{TimesNet}
& \scalebox{0.75}{\textbf{Ours}} \\

\midrule
\multirow{10}{*}{\rotatebox{90}{{\textbf{ETTh2}}}}
& \scalebox{0.75}{MAE@0\%} & \scalebox{0.75}{0.096} & \scalebox{0.75}{0.095} & \scalebox{0.75}{0.098} & \scalebox{0.75}{0.096} & \scalebox{0.75}{0.098} & \scalebox{0.75}{0.096} & \scalebox{0.75}{0.113} & \scalebox{0.75}{0.197} & \scalebox{0.75}{0.101} & \scalebox{0.75}{0.097} & \scalebox{0.75}{\myfirst{0.094}} \\
& \scalebox{0.75}{MSE@0\%} & \scalebox{0.75}{0.024} & \scalebox{0.75}{0.024} & \scalebox{0.75}{0.025} & \scalebox{0.75}{0.024} & \scalebox{0.75}{0.025} & \scalebox{0.75}{0.025} & \scalebox{0.75}{0.033} & \scalebox{0.75}{0.074} & \scalebox{0.75}{0.026} & \scalebox{0.75}{0.025} & \scalebox{0.75}{\myfirst{0.023}} \\
\cmidrule(lr){2-2} \cmidrule(lr){3-13}
& \scalebox{0.75}{MAE@20\%} & \scalebox{0.75}{0.115} & \scalebox{0.75}{0.116} & \scalebox{0.75}{0.115} & \scalebox{0.75}{0.109} & \scalebox{0.75}{0.142} & \scalebox{0.75}{0.116} & \scalebox{0.75}{0.255} & \scalebox{0.75}{0.244} & \scalebox{0.75}{0.122} & \scalebox{0.75}{0.121} & \scalebox{0.75}{\myfirst{0.107}} \\
& \scalebox{0.75}{MSE@20\%} & \scalebox{0.75}{0.031} & \scalebox{0.75}{0.030} & \scalebox{0.75}{0.033} & \scalebox{0.75}{0.030} & \scalebox{0.75}{0.040} & \scalebox{0.75}{0.033} & \scalebox{0.75}{0.192} & \scalebox{0.75}{0.105} & \scalebox{0.75}{0.034} & \scalebox{0.75}{0.033} & \scalebox{0.75}{\myfirst{0.028}} \\
\cmidrule(lr){2-2} \cmidrule(lr){3-13}
& \scalebox{0.75}{MAE@40\%} & \scalebox{0.75}{0.148} & \scalebox{0.75}{0.147} & \scalebox{0.75}{0.298} & \scalebox{0.75}{0.215} & \scalebox{0.75}{0.181} & \scalebox{0.75}{0.289} & \scalebox{0.75}{0.404} & \scalebox{0.75}{0.276} & \scalebox{0.75}{0.140} & \scalebox{0.75}{0.153} & \scalebox{0.75}{\myfirst{0.131}} \\
& \scalebox{0.75}{MSE@40\%} & \scalebox{0.75}{0.047} & \scalebox{0.75}{0.045} & \scalebox{0.75}{0.180} & \scalebox{0.75}{0.125} & \scalebox{0.75}{0.065} & \scalebox{0.75}{0.173} & \scalebox{0.75}{0.399} & \scalebox{0.75}{0.134} & \scalebox{0.75}{0.049} & \scalebox{0.75}{0.048} & \scalebox{0.75}{\myfirst{0.039}} \\
\cmidrule(lr){2-2} \cmidrule(lr){3-13}
& \scalebox{0.75}{MAE@60\%} & \scalebox{0.75}{0.183} & \scalebox{0.75}{0.202} & \scalebox{0.75}{0.270} & \scalebox{0.75}{0.216} & \scalebox{0.75}{0.253} & \scalebox{0.75}{0.243} & \scalebox{0.75}{0.618} & \scalebox{0.75}{0.347} & \scalebox{0.75}{0.173} & \scalebox{0.75}{0.209} & \scalebox{0.75}{\myfirst{0.154}} \\
& \scalebox{0.75}{MSE@60\%} & \scalebox{0.75}{0.072} & \scalebox{0.75}{0.086} & \scalebox{0.75}{0.170} & \scalebox{0.75}{0.121} & \scalebox{0.75}{0.135} & \scalebox{0.75}{0.133} & \scalebox{0.75}{0.840} & \scalebox{0.75}{0.210} & \scalebox{0.75}{0.061} & \scalebox{0.75}{0.092} & \scalebox{0.75}{\myfirst{0.055}} \\
\cmidrule(lr){2-2} \cmidrule(lr){3-13}
& \scalebox{0.75}{MAE@70\%} & \scalebox{0.75}{0.205} & \scalebox{0.75}{0.218} & \scalebox{0.75}{0.218} & \scalebox{0.75}{0.186} & \scalebox{0.75}{0.325} & \scalebox{0.75}{0.196} & \scalebox{0.75}{0.798} & \scalebox{0.75}{0.413} & \scalebox{0.75}{0.174} & \scalebox{0.75}{0.246} & \scalebox{0.75}{\myfirst{0.172}} \\
& \scalebox{0.75}{MSE@70\%} & \scalebox{0.75}{0.093} & \scalebox{0.75}{0.104} & \scalebox{0.75}{0.111} & \scalebox{0.75}{0.083} & \scalebox{0.75}{0.228} & \scalebox{0.75}{0.085} & \scalebox{0.75}{1.303} & \scalebox{0.75}{0.303} & \scalebox{0.75}{0.071} & \scalebox{0.75}{0.132} & \scalebox{0.75}{\myfirst{0.069}} \\

\midrule

\multirow{10}{*}{\rotatebox{90}{{\textbf{ETTm1}}}}
& \scalebox{0.75}{MAE@0\%} & \scalebox{0.75}{0.2265} & \scalebox{0.75}{0.2405} & \scalebox{0.75}{0.2309} & \scalebox{0.75}{0.2379} & \scalebox{0.75}{0.2772} & \scalebox{0.75}{0.2423} & \scalebox{0.75}{0.4759} & \scalebox{0.75}{0.3331} & \scalebox{0.75}{0.2999} & \scalebox{0.75}{0.2100} & \scalebox{0.75}{\myfirst{0.2000}} \\
& \scalebox{0.75}{MSE@0\%} & \scalebox{0.75}{0.2302} & \scalebox{0.75}{0.2741} & \scalebox{0.75}{0.2514} & \scalebox{0.75}{0.2648} & \scalebox{0.75}{0.3504} & \scalebox{0.75}{0.2782} & \scalebox{0.75}{0.8854} & \scalebox{0.75}{0.4218} & \scalebox{0.75}{0.2682} & \scalebox{0.75}{0.2018} & \scalebox{0.75}{\myfirst{0.2008}} \\
\cmidrule(lr){2-2} \cmidrule(lr){3-13}
& \scalebox{0.75}{MAE@20\%} & \scalebox{0.75}{0.2616} & \scalebox{0.75}{0.2900} & \scalebox{0.75}{0.2909} & \scalebox{0.75}{0.2779} & \scalebox{0.75}{0.3711} & \scalebox{0.75}{0.2971} & \scalebox{0.75}{0.5198} & \scalebox{0.75}{0.3809} & \scalebox{0.75}{0.3271} & \scalebox{0.75}{0.2584} & \scalebox{0.75}{\myfirst{0.2368}} \\
& \scalebox{0.75}{MSE@20\%} & \scalebox{0.75}{0.3068} & \scalebox{0.75}{0.3593} & \scalebox{0.75}{0.3751} & \scalebox{0.75}{0.3486} & \scalebox{0.75}{0.5499} & \scalebox{0.75}{0.3864} & \scalebox{0.75}{0.9345} & \scalebox{0.75}{0.5177} & \scalebox{0.75}{0.3282} & \scalebox{0.75}{0.2992} & \scalebox{0.75}{\myfirst{0.2586}} \\
\cmidrule(lr){2-2} \cmidrule(lr){3-13}
& \scalebox{0.75}{MAE@40\%} & \scalebox{0.75}{0.3005} & \scalebox{0.75}{0.3308} & \scalebox{0.75}{0.3593} & \scalebox{0.75}{0.3456} & \scalebox{0.75}{0.4418} & \scalebox{0.75}{0.3592} & \scalebox{0.75}{0.5787} & \scalebox{0.75}{0.4391} & \scalebox{0.75}{0.3571} & \scalebox{0.75}{0.3040} & \scalebox{0.75}{\myfirst{0.2734}} \\
& \scalebox{0.75}{MSE@40\%} & \scalebox{0.75}{0.4007} & \scalebox{0.75}{0.4645} & \scalebox{0.75}{0.4851} & \scalebox{0.75}{0.4671} & \scalebox{0.75}{0.7412} & \scalebox{0.75}{0.4801} & \scalebox{0.75}{1.0499} & \scalebox{0.75}{0.6643} & \scalebox{0.75}{0.4749} & \scalebox{0.75}{0.4093} & \scalebox{0.75}{\myfirst{0.3382}} \\
\cmidrule(lr){2-2} \cmidrule(lr){3-13}
& \scalebox{0.75}{MAE@60\%} & \scalebox{0.75}{0.3678} & \scalebox{0.75}{0.4171} & \scalebox{0.75}{0.4365} & \scalebox{0.75}{0.4091} & \scalebox{0.75}{0.5551} & \scalebox{0.75}{0.4191} & \scalebox{0.75}{0.7038} & \scalebox{0.75}{0.5457} & \scalebox{0.75}{0.4315} & \scalebox{0.75}{0.3993} & \scalebox{0.75}{\myfirst{0.3414}} \\
& \scalebox{0.75}{MSE@60\%} & \scalebox{0.75}{0.5960} & \scalebox{0.75}{0.7328} & \scalebox{0.75}{0.7580} & \scalebox{0.75}{0.6769} & \scalebox{0.75}{1.1067} & \scalebox{0.75}{0.7022} & \scalebox{0.75}{1.3801} & \scalebox{0.75}{0.9573} & \scalebox{0.75}{0.6978} & \scalebox{0.75}{0.6525} & \scalebox{0.75}{\myfirst{0.5200}} \\
\cmidrule(lr){2-2} \cmidrule(lr){3-13}
& \scalebox{0.75}{MAE@70\%} & \scalebox{0.75}{0.4266} & \scalebox{0.75}{0.4863} & \scalebox{0.75}{0.5034} & \scalebox{0.75}{0.4743} & \scalebox{0.75}{0.6508} & \scalebox{0.75}{0.4816} & \scalebox{0.75}{0.8287} & \scalebox{0.75}{0.6278} & \scalebox{0.75}{0.4845} & \scalebox{0.75}{0.4828} & \scalebox{0.75}{\myfirst{0.4013}} \\
& \scalebox{0.75}{MSE@70\%} & \scalebox{0.75}{0.8006} & \scalebox{0.75}{0.9928} & \scalebox{0.75}{1.0663} & \scalebox{0.75}{0.9325} & \scalebox{0.75}{1.4721} & \scalebox{0.75}{0.9837} & \scalebox{0.75}{1.7733} & \scalebox{0.75}{1.2710} & \scalebox{0.75}{0.9035} & \scalebox{0.75}{0.8988} & \scalebox{0.75}{\myfirst{0.6956}} \\

\midrule

\multirow{10}{*}{\rotatebox{90}{{\textbf{ETTm2}}}}
& \scalebox{0.75}{MAE@0\%} & \scalebox{0.75}{0.0802} & \scalebox{0.75}{0.0821} & \scalebox{0.75}{0.0781} & \scalebox{0.75}{0.0824} & \scalebox{0.75}{0.0918} & \scalebox{0.75}{0.0828} & \scalebox{0.75}{0.1313} & \scalebox{0.75}{0.1327} & \scalebox{0.75}{0.0813} & \scalebox{0.75}{0.0782} & \scalebox{0.75}{\myfirst{0.0746}} \\
& \scalebox{0.75}{MSE@0\%} & \scalebox{0.75}{0.0175} & \scalebox{0.75}{0.0182} & \scalebox{0.75}{0.0161} & \scalebox{0.75}{0.0182} & \scalebox{0.75}{0.0217} & \scalebox{0.75}{0.0187} & \scalebox{0.75}{0.0410} & \scalebox{0.75}{0.0363} & \scalebox{0.75}{0.0160} & \scalebox{0.75}{0.0165} & \scalebox{0.75}{\myfirst{0.0152}} \\
\cmidrule(lr){2-2} \cmidrule(lr){3-13}
& \scalebox{0.75}{MAE@20\%} & \scalebox{0.75}{0.0956} & \scalebox{0.75}{0.1059} & \scalebox{0.75}{0.1020} & \scalebox{0.75}{0.1005} & \scalebox{0.75}{0.1405} & \scalebox{0.75}{0.1024} & \scalebox{0.75}{0.2687} & \scalebox{0.75}{0.1760} & \scalebox{0.75}{0.0888} & \scalebox{0.75}{0.1087} & \scalebox{0.75}{\myfirst{0.0885}} \\
& \scalebox{0.75}{MSE@20\%} & \scalebox{0.75}{0.0231} & \scalebox{0.75}{0.0266} & \scalebox{0.75}{0.0284} & \scalebox{0.75}{0.0251} & \scalebox{0.75}{0.0391} & \scalebox{0.75}{0.0278} & \scalebox{0.75}{0.1976} & \scalebox{0.75}{0.0591} & \scalebox{0.75}{0.0222} & \scalebox{0.75}{0.0260} & \scalebox{0.75}{\myfirst{0.0203}} \\
\cmidrule(lr){2-2} \cmidrule(lr){3-13}
& \scalebox{0.75}{MAE@40\%} & \scalebox{0.75}{0.1159} & \scalebox{0.75}{0.1355} & \scalebox{0.75}{0.2673} & \scalebox{0.75}{0.2041} & \scalebox{0.75}{0.1743} & \scalebox{0.75}{0.2441} & \scalebox{0.75}{0.4131} & \scalebox{0.75}{0.1984} & \scalebox{0.75}{0.1088} & \scalebox{0.75}{0.1417} & \scalebox{0.75}{\myfirst{0.1037}} \\
& \scalebox{0.75}{MSE@40\%} & \scalebox{0.75}{0.0305} & \scalebox{0.75}{0.0394} & \scalebox{0.75}{0.1688} & \scalebox{0.75}{0.1210} & \scalebox{0.75}{0.0601} & \scalebox{0.75}{0.1490} & \scalebox{0.75}{0.4003} & \scalebox{0.75}{0.0765} & \scalebox{0.75}{0.0267} & \scalebox{0.75}{0.0406} & \scalebox{0.75}{\myfirst{0.0258}} \\
\cmidrule(lr){2-2} \cmidrule(lr){3-13}
& \scalebox{0.75}{MAE@60\%} & \scalebox{0.75}{0.1432} & \scalebox{0.75}{0.1790} & \scalebox{0.75}{0.2651} & \scalebox{0.75}{0.2425} & \scalebox{0.75}{0.2412} & \scalebox{0.75}{0.2398} & \scalebox{0.75}{0.6221} & \scalebox{0.75}{0.2479} & \scalebox{0.75}{0.1360} & \scalebox{0.75}{0.1986} & \scalebox{0.75}{\myfirst{0.1266}} \\
& \scalebox{0.75}{MSE@60\%} & \scalebox{0.75}{0.0477} & \scalebox{0.75}{0.0685} & \scalebox{0.75}{0.1671} & \scalebox{0.75}{0.1423} & \scalebox{0.75}{0.1215} & \scalebox{0.75}{0.1448} & \scalebox{0.75}{0.8293} & \scalebox{0.75}{0.1194} & \scalebox{0.75}{0.0424} & \scalebox{0.75}{0.0811} & \scalebox{0.75}{\myfirst{0.0384}} \\
\cmidrule(lr){2-2} \cmidrule(lr){3-13}
& \scalebox{0.75}{MAE@70\%} & \scalebox{0.75}{0.1650} & \scalebox{0.75}{0.1899} & \scalebox{0.75}{0.1905} & \scalebox{0.75}{0.1803} & \scalebox{0.75}{0.3104} & \scalebox{0.75}{0.1728} & \scalebox{0.75}{0.8024} & \scalebox{0.75}{0.3130} & \scalebox{0.75}{0.1602} & \scalebox{0.75}{0.2317} & \scalebox{0.75}{\myfirst{0.1467}} \\
& \scalebox{0.75}{MSE@70\%} & \scalebox{0.75}{0.0619} & \scalebox{0.75}{0.0808} & \scalebox{0.75}{0.0861} & \scalebox{0.75}{0.0779} & \scalebox{0.75}{0.2056} & \scalebox{0.75}{0.0717} & \scalebox{0.75}{1.2968} & \scalebox{0.75}{0.1857} & \scalebox{0.75}{0.0569} & \scalebox{0.75}{0.1142} & \scalebox{0.75}{\myfirst{0.0520}} \\

\midrule

\multirow{10}{*}{\rotatebox{90}{{\textbf{Weather}}}}
& \scalebox{0.75}{MAE@0\%} & \scalebox{0.75}{0.031} & \scalebox{0.75}{0.030} & \scalebox{0.75}{0.030} & \scalebox{0.75}{0.031} & \scalebox{0.75}{0.034} & \scalebox{0.75}{0.031} & \scalebox{0.75}{0.051} & \scalebox{0.75}{0.051} & \scalebox{0.75}{0.035} & \scalebox{0.75}{0.028} & \scalebox{0.75}{\myfirst{0.028}} \\
& \scalebox{0.75}{MSE@0\%} & \scalebox{0.75}{0.021} & \scalebox{0.75}{0.019} & \scalebox{0.75}{0.018} & \scalebox{0.75}{0.019} & \scalebox{0.75}{0.023} & \scalebox{0.75}{0.020} & \scalebox{0.75}{0.042} & \scalebox{0.75}{0.029} & \scalebox{0.75}{0.020} & \scalebox{0.75}{0.016} & \scalebox{0.75}{\myfirst{0.016}} \\
\cmidrule(lr){2-2} \cmidrule(lr){3-13}
& \scalebox{0.75}{MAE@20\%} & \scalebox{0.75}{0.042} & \scalebox{0.75}{0.050} & \scalebox{0.75}{0.044} & \scalebox{0.75}{0.048} & \scalebox{0.75}{0.099} & \scalebox{0.75}{0.046} & \scalebox{0.75}{0.173} & \scalebox{0.75}{0.073} & \scalebox{0.75}{0.041} & \scalebox{0.75}{0.062} & \scalebox{0.75}{\myfirst{0.038}} \\
& \scalebox{0.75}{MSE@20\%} & \scalebox{0.75}{0.028} & \scalebox{0.75}{0.025} & \scalebox{0.75}{0.031} & \scalebox{0.75}{0.027} & \scalebox{0.75}{0.045} & \scalebox{0.75}{0.032} & \scalebox{0.75}{0.220} & \scalebox{0.75}{0.036} & \scalebox{0.75}{0.028} & \scalebox{0.75}{0.033} & \scalebox{0.75}{\myfirst{0.025}} \\
\cmidrule(lr){2-2} \cmidrule(lr){3-13}
& \scalebox{0.75}{MAE@40\%} & \scalebox{0.75}{0.050} & \scalebox{0.75}{0.074} & \scalebox{0.75}{0.152} & \scalebox{0.75}{0.150} & \scalebox{0.75}{0.135} & \scalebox{0.75}{0.157} & \scalebox{0.75}{0.300} & \scalebox{0.75}{0.095} & \scalebox{0.75}{0.051} & \scalebox{0.75}{0.088} & \scalebox{0.75}{\myfirst{0.050}} \\
& \scalebox{0.75}{MSE@40\%} & \scalebox{0.75}{0.037} & \scalebox{0.75}{0.036} & \scalebox{0.75}{0.167} & \scalebox{0.75}{0.156} & \scalebox{0.75}{0.078} & \scalebox{0.75}{0.140} & \scalebox{0.75}{0.500} & \scalebox{0.75}{0.060} & \scalebox{0.75}{0.033} & \scalebox{0.75}{0.047} & \scalebox{0.75}{\myfirst{0.033}} \\
\cmidrule(lr){2-2} \cmidrule(lr){3-13}
& \scalebox{0.75}{MAE@60\%} & \scalebox{0.75}{0.066} & \scalebox{0.75}{0.102} & \scalebox{0.75}{0.175} & \scalebox{0.75}{0.169} & \scalebox{0.75}{0.197} & \scalebox{0.75}{0.168} & \scalebox{0.75}{0.466} & \scalebox{0.75}{0.150} & \scalebox{0.75}{0.066} & \scalebox{0.75}{0.128} & \scalebox{0.75}{\myfirst{0.062}} \\
& \scalebox{0.75}{MSE@60\%} & \scalebox{0.75}{0.061} & \scalebox{0.75}{0.062} & \scalebox{0.75}{0.223} & \scalebox{0.75}{0.169} & \scalebox{0.75}{0.178} & \scalebox{0.75}{0.163} & \scalebox{0.75}{1.099} & \scalebox{0.75}{0.150} & \scalebox{0.75}{0.064} & \scalebox{0.75}{0.089} & \scalebox{0.75}{\myfirst{0.057}} \\
\cmidrule(lr){2-2} \cmidrule(lr){3-13}
& \scalebox{0.75}{MAE@70\%} & \scalebox{0.75}{0.076} & \scalebox{0.75}{0.121} & \scalebox{0.75}{0.101} & \scalebox{0.75}{0.118} & \scalebox{0.75}{0.254} & \scalebox{0.75}{0.100} & \scalebox{0.75}{0.586} & \scalebox{0.75}{0.204} & \scalebox{0.75}{0.076} & \scalebox{0.75}{0.166} & \scalebox{0.75}{\myfirst{0.070}} \\
& \scalebox{0.75}{MSE@70\%} & \scalebox{0.75}{0.081} & \scalebox{0.75}{0.084} & \scalebox{0.75}{0.138} & \scalebox{0.75}{0.128} & \scalebox{0.75}{0.318} & \scalebox{0.75}{0.120} & \scalebox{0.75}{1.707} & \scalebox{0.75}{0.284} & \scalebox{0.75}{0.083} & \scalebox{0.75}{0.154} & \scalebox{0.75}{\myfirst{0.075}} \\

\midrule

\multirow{10}{*}{\rotatebox{90}{{\textbf{Exchange}}}}
& \scalebox{0.75}{MAE@0\%} & \scalebox{0.75}{0.0185} & \scalebox{0.75}{0.0187} & \scalebox{0.75}{0.0190} & \scalebox{0.75}{0.0186} & \scalebox{0.75}{0.0205} & \scalebox{0.75}{0.0187} & \scalebox{0.75}{0.0264} & \scalebox{0.75}{0.3223} & \scalebox{0.75}{0.0344} & \scalebox{0.75}{0.0195} & \scalebox{0.75}{\myfirst{0.0184}} \\
& \scalebox{0.75}{MSE@0\%} & \scalebox{0.75}{0.0010} & \scalebox{0.75}{0.0010} & \scalebox{0.75}{0.0010} & \scalebox{0.75}{0.0010} & \scalebox{0.75}{0.0011} & \scalebox{0.75}{0.0010} & \scalebox{0.75}{0.0017} & \scalebox{0.75}{0.1724} & \scalebox{0.75}{0.0031} & \scalebox{0.75}{0.0011} & \scalebox{0.75}{\myfirst{0.0009}} \\
\cmidrule(lr){2-2} \cmidrule(lr){3-13}
& \scalebox{0.75}{MAE@20\%} & \scalebox{0.75}{0.0292} & \scalebox{0.75}{0.0324} & \scalebox{0.75}{0.0288} & \scalebox{0.75}{0.0273} & \scalebox{0.75}{0.0795} & \scalebox{0.75}{0.0227} & \scalebox{0.75}{0.1992} & \scalebox{0.75}{0.2969} & \scalebox{0.75}{0.0364} & \scalebox{0.75}{0.0458} & \scalebox{0.75}{\myfirst{0.0217}} \\
& \scalebox{0.75}{MSE@20\%} & \scalebox{0.75}{0.0031} & \scalebox{0.75}{0.0024} & \scalebox{0.75}{0.0024} & \scalebox{0.75}{0.0019} & \scalebox{0.75}{0.0126} & \scalebox{0.75}{0.0016} & \scalebox{0.75}{0.2089} & \scalebox{0.75}{0.1592} & \scalebox{0.75}{0.0029} & \scalebox{0.75}{0.0041} & \scalebox{0.75}{\myfirst{0.0016}} \\
\cmidrule(lr){2-2} \cmidrule(lr){3-13}
& \scalebox{0.75}{MAE@40\%} & \scalebox{0.75}{0.0629} & \scalebox{0.75}{0.0783} & \scalebox{0.75}{0.1899} & \scalebox{0.75}{0.1416} & \scalebox{0.75}{0.1469} & \scalebox{0.75}{0.1408} & \scalebox{0.75}{0.4358} & \scalebox{0.75}{0.3326} & \scalebox{0.75}{0.0431} & \scalebox{0.75}{0.0786} & \scalebox{0.75}{\myfirst{0.0253}} \\
& \scalebox{0.75}{MSE@40\%} & \scalebox{0.75}{0.0129} & \scalebox{0.75}{0.0127} & \scalebox{0.75}{0.0770} & \scalebox{0.75}{0.0884} & \scalebox{0.75}{0.0429} & \scalebox{0.75}{0.1047} & \scalebox{0.75}{0.5718} & \scalebox{0.75}{0.2146} & \scalebox{0.75}{0.0034} & \scalebox{0.75}{0.0130} & \scalebox{0.75}{\myfirst{0.0015}} \\
\cmidrule(lr){2-2} \cmidrule(lr){3-13}
& \scalebox{0.75}{MAE@60\%} & \scalebox{0.75}{0.1165} & \scalebox{0.75}{0.1364} & \scalebox{0.75}{0.2439} & \scalebox{0.75}{0.1978} & \scalebox{0.75}{0.2599} & \scalebox{0.75}{0.1283} & \scalebox{0.75}{0.7911} & \scalebox{0.75}{0.4334} & \scalebox{0.75}{0.0608} & \scalebox{0.75}{0.1357} & \scalebox{0.75}{\myfirst{0.0363}} \\
& \scalebox{0.75}{MSE@60\%} & \scalebox{0.75}{0.0363} & \scalebox{0.75}{0.0405} & \scalebox{0.75}{0.1060} & \scalebox{0.75}{0.1129} & \scalebox{0.75}{0.1394} & \scalebox{0.75}{0.0831} & \scalebox{0.75}{1.3714} & \scalebox{0.75}{0.3814} & \scalebox{0.75}{0.0071} & \scalebox{0.75}{0.0404} & \scalebox{0.75}{\myfirst{0.0031}} \\
\cmidrule(lr){2-2} \cmidrule(lr){3-13}
& \scalebox{0.75}{MAE@70\%} & \scalebox{0.75}{0.1418} & \scalebox{0.75}{0.1836} & \scalebox{0.75}{0.2865} & \scalebox{0.75}{0.1321} & \scalebox{0.75}{0.3778} & \scalebox{0.75}{0.0718} & \scalebox{0.75}{1.0766} & \scalebox{0.75}{0.5412} & \scalebox{0.75}{0.0846} & \scalebox{0.75}{0.1991} & \scalebox{0.75}{\myfirst{0.0517}} \\
& \scalebox{0.75}{MSE@70\%} & \scalebox{0.75}{0.0490} & \scalebox{0.75}{0.0697} & \scalebox{0.75}{0.1413} & \scalebox{0.75}{0.0581} & \scalebox{0.75}{0.3011} & \scalebox{0.75}{0.0169} & \scalebox{0.75}{2.2850} & \scalebox{0.75}{0.5912} & \scalebox{0.75}{0.0128} & \scalebox{0.75}{0.0881} & \scalebox{0.75}{\myfirst{0.0058}} \\

\midrule

\multirow{10}{*}{\rotatebox{90}{{\textbf{BeijingAir}}}}
& \scalebox{0.75}{MAE@0\%} & \scalebox{0.75}{0.2552} & \scalebox{0.75}{0.2571} & \scalebox{0.75}{0.2606} & \scalebox{0.75}{0.2609} & \scalebox{0.75}{0.2707} & \scalebox{0.75}{0.2601} & \scalebox{0.75}{0.3431} & \scalebox{0.75}{0.3167} & \scalebox{0.75}{0.2533} & \scalebox{0.75}{0.2583} & \scalebox{0.75}{\myfirst{0.2526}} \\
& \scalebox{0.75}{MSE@0\%} & \scalebox{0.75}{0.3026} & \scalebox{0.75}{0.3085} & \scalebox{0.75}{0.3156} & \scalebox{0.75}{0.3008} & \scalebox{0.75}{0.3263} & \scalebox{0.75}{0.3102} & \scalebox{0.75}{0.5141} & \scalebox{0.75}{0.3779} & \scalebox{0.75}{0.3189} & \scalebox{0.75}{0.2939} & \scalebox{0.75}{\myfirst{0.2929}} \\
\cmidrule(lr){2-2} \cmidrule(lr){3-13}
& \scalebox{0.75}{MAE@20\%} & \scalebox{0.75}{0.2849} & \scalebox{0.75}{0.2896} & \scalebox{0.75}{0.2923} & \scalebox{0.75}{0.2866} & \scalebox{0.75}{0.3204} & \scalebox{0.75}{0.2949} & \scalebox{0.75}{0.4035} & \scalebox{0.75}{0.3478} & \scalebox{0.75}{0.2770} & \scalebox{0.75}{0.2922} & \scalebox{0.75}{\myfirst{0.2758}} \\
& \scalebox{0.75}{MSE@20\%} & \scalebox{0.75}{0.3467} & \scalebox{0.75}{0.3651} & \scalebox{0.75}{0.3655} & \scalebox{0.75}{0.3421} & \scalebox{0.75}{0.3853} & \scalebox{0.75}{0.3603} & \scalebox{0.75}{0.6010} & \scalebox{0.75}{0.4196} & \scalebox{0.75}{0.3496} & \scalebox{0.75}{0.3488} & \scalebox{0.75}{\myfirst{0.3317}} \\
\cmidrule(lr){2-2} \cmidrule(lr){3-13}
& \scalebox{0.75}{MAE@40\%} & \scalebox{0.75}{0.3188} & \scalebox{0.75}{0.3312} & \scalebox{0.75}{0.3545} & \scalebox{0.75}{0.3278} & \scalebox{0.75}{0.3620} & \scalebox{0.75}{0.3480} & \scalebox{0.75}{0.4714} & \scalebox{0.75}{0.3950} & \scalebox{0.75}{0.3297} & \scalebox{0.75}{0.3322} & \scalebox{0.75}{\myfirst{0.3106}} \\
& \scalebox{0.75}{MSE@40\%} & \scalebox{0.75}{0.4285} & \scalebox{0.75}{0.4601} & \scalebox{0.75}{0.4820} & \scalebox{0.75}{0.4305} & \scalebox{0.75}{0.4760} & \scalebox{0.75}{0.4689} & \scalebox{0.75}{0.7584} & \scalebox{0.75}{0.5474} & \scalebox{0.75}{0.4645} & \scalebox{0.75}{0.4422} & \scalebox{0.75}{\myfirst{0.4120}} \\
\cmidrule(lr){2-2} \cmidrule(lr){3-13}
& \scalebox{0.75}{MAE@60\%} & \scalebox{0.75}{0.3860} & \scalebox{0.75}{0.4011} & \scalebox{0.75}{0.4124} & \scalebox{0.75}{0.3976} & \scalebox{0.75}{0.4362} & \scalebox{0.75}{0.4142} & \scalebox{0.75}{0.5930} & \scalebox{0.75}{0.4750} & \scalebox{0.75}{0.4006} & \scalebox{0.75}{0.4017} & \scalebox{0.75}{\myfirst{0.3728}} \\
& \scalebox{0.75}{MSE@60\%} & \scalebox{0.75}{0.5939} & \scalebox{0.75}{0.6503} & \scalebox{0.75}{0.6573} & \scalebox{0.75}{0.6458} & \scalebox{0.75}{0.6799} & \scalebox{0.75}{0.6624} & \scalebox{0.75}{1.1378} & \scalebox{0.75}{0.7803} & \scalebox{0.75}{0.6110} & \scalebox{0.75}{0.6275} & \scalebox{0.75}{\myfirst{0.5862}} \\
\cmidrule(lr){2-2} \cmidrule(lr){3-13}
& \scalebox{0.75}{MAE@70\%} & \scalebox{0.75}{0.4412} & \scalebox{0.75}{0.4632} & \scalebox{0.75}{0.4691} & \scalebox{0.75}{0.4613} & \scalebox{0.75}{0.5022} & \scalebox{0.75}{0.4673} & \scalebox{0.75}{0.7057} & \scalebox{0.75}{0.5456} & \scalebox{0.75}{0.4549} & \scalebox{0.75}{0.4789} & \scalebox{0.75}{\myfirst{0.4337}} \\
& \scalebox{0.75}{MSE@70\%} & \scalebox{0.75}{\myfirst{0.7595}} & \scalebox{0.75}{0.8730} & \scalebox{0.75}{0.8874} & \scalebox{0.75}{0.8540} & \scalebox{0.75}{0.9358} & \scalebox{0.75}{0.8745} & \scalebox{0.75}{1.5761} & \scalebox{0.75}{1.0151} & \scalebox{0.75}{0.8327} & \scalebox{0.75}{0.9234} & \scalebox{0.75}{0.8010} \\
\bottomrule
\end{tabular}
}
\end{scriptsize}
\end{table*}




\end{document}